\documentclass{article}

\usepackage{times}
\usepackage{graphicx} 
\usepackage{subfigure} 
\usepackage{microtype}

\usepackage{url}

\usepackage[round]{natbib}
\usepackage{amsmath,amssymb}

\usepackage{xspace}

\usepackage{natbib}
\usepackage{bibentry}
\nobibliography*
\usepackage{hanging}

\usepackage{algorithm}
\usepackage{algorithmic}

\usepackage[usenames,dvipsnames]{xcolor}

\usepackage{booktabs}

\newcommand{\obsI}{\hat{I}}

\newcommand{\INDLMH}{INF-INDMH\xspace}
\newcommand{\MIXLMH}{INF-MH\xspace}
\newcommand{\INFBMHWG}{INF-BMHWG\xspace}
\newcommand{\REGMH}{REG-MH\xspace}
\newcommand{\REGAMH}{REG-MH\xspace}
\newcommand{\MIXLMHRF}{INF-RFMH\xspace}
\newcommand{\MH}{MH\xspace}
\newcommand{\MHWG}{MHWG\xspace}
\newcommand{\PT}{PT\xspace}
\newcommand{\BMHWG}{BMHWG\xspace}

\usepackage[accepted]{icml2014_arxiv}

\icmltitlerunning{The Informed Sampler: A Discriminative Approach to
  Bayesian Inference in Computer Vision}

\begin{document} 

\twocolumn[
\icmltitle{The Informed Sampler:~\\ A Discriminative Approach to Bayesian Inference in~\\
Generative Computer Vision Models}

\icmlauthor{Varun Jampani}{varun.jampani@tuebingen.mpg.de}
\icmladdress{Max Planck Institute for Intelligent Systems, Spemannstra{\ss}e 41, 72076, T{\"u}bingen, Germany}

\icmlauthor{Sebastian Nowozin}{sebastian.nowozin@microsoft.com}
\icmladdress{Microsoft Research Cambridge, 21 Station Road, Cambridge, CB1 2FB, United Kingdom}

\icmlauthor{Matthew Loper}{mloper@tuebingen.mpg.de}
\icmladdress{Max Planck Institute for Intelligent Systems, Spemannstra{\ss}e 41, 72076, T{\"u}bingen, Germany}

\icmlauthor{Peter V. Gehler}{pgehler@tuebingen.mpg.de}
\icmladdress{Max Planck Institute for Intelligent Systems, Spemannstra{\ss}e 41, 72076, T{\"u}bingen, Germany}

          \icmlkeywords{probabilistic models, markov chain monte
            carlo, inverse graphics, computer vision, generative
            models}

\vskip 0.2in
]

\begin{abstract}
Computer vision is hard because of a large variability in lighting, shape,
and texture; in addition the image signal is non-additive due to occlusion.
Generative models promised to account for this variability by accurately
modelling the image formation process as a function of latent variables with
prior beliefs.
Bayesian posterior inference could then, in principle, explain the observation.
While intuitively appealing, generative models for computer vision have
largely failed to deliver on that promise due to the difficulty of posterior
inference.
As a result the community has favoured efficient discriminative approaches.
We still believe in the usefulness of generative models in computer vision,
but argue that we need to leverage existing discriminative or even heuristic
computer vision methods.
We implement this idea in a principled way with an \emph{informed sampler} and
in careful experiments demonstrate it on challenging generative models which contain
renderer programs as their components.
We concentrate on the problem of inverting an existing graphics rendering engine, an approach
that can be understood as ``Inverse Graphics".
The informed sampler, using simple discriminative proposals based on existing
computer vision technology, achieves significant improvements of inference.
\end{abstract}

\vspace{-0.5cm}

\section{Introduction}
\label{sec:introduction}
A conceptually elegant view on computer vision is to consider a generative
model of the physical image formation process.
The observed image becomes a function of unobserved variables of interest (for
example presence and positions of objects) and nuisance variables (for example
light sources, shadows).
When building such a generative model, we can think of a scene description
$\theta$ that produces an image $I=G(\theta)$ using a deterministic rendering
engine $G$, or more generally, results in a distribution over images, $p(I|\theta)$.
Given an image observation $\obsI$ and a prior over scenes $p(\theta)$ we can
then perform Bayesian inference to obtain updated beliefs $p(\theta | \obsI)$.
This view was advocated since the late 1970'ies~\cite{horn1977imageintensities,grenander1976patterntheory,zhu1997learning,mumford2010patterntheory,mansinghka2013approximate,yuille2006vision}.
%

Now, 30 years later, we would argue that the generative approach has largely failed to
deliver on its promise.
The few successes of the idea have been in limited settings.
In the successful examples, either
the generative model was restricted to few high-level latent
variables, e.g.~\cite{oliver2000humaninteractions},
or restricted to a set of image transformations in a fixed reference
frame, e.g.~\cite{black2000imageappearance},
or it modelled only a limited aspect such as
object shape masks~\cite{eslami2012shapeboltzmann},
or, in the worst case, the generative model was merely used to generate
training data for a discriminative model~\cite{shotton2011kinect}.
With all its intuitive appeal, its beauty and simplicity, it is fair to say
that the track record of generative models in computer vision is poor.
As a result, the field of computer vision is now dominated by
efficient but data-hungry discriminative models,
the use of empirical risk minimization for learning,
and energy minimization on heuristic objective functions for inference.

\begin{figure}[t]
\begin{center}
\centerline{\includegraphics[width=0.7\columnwidth]{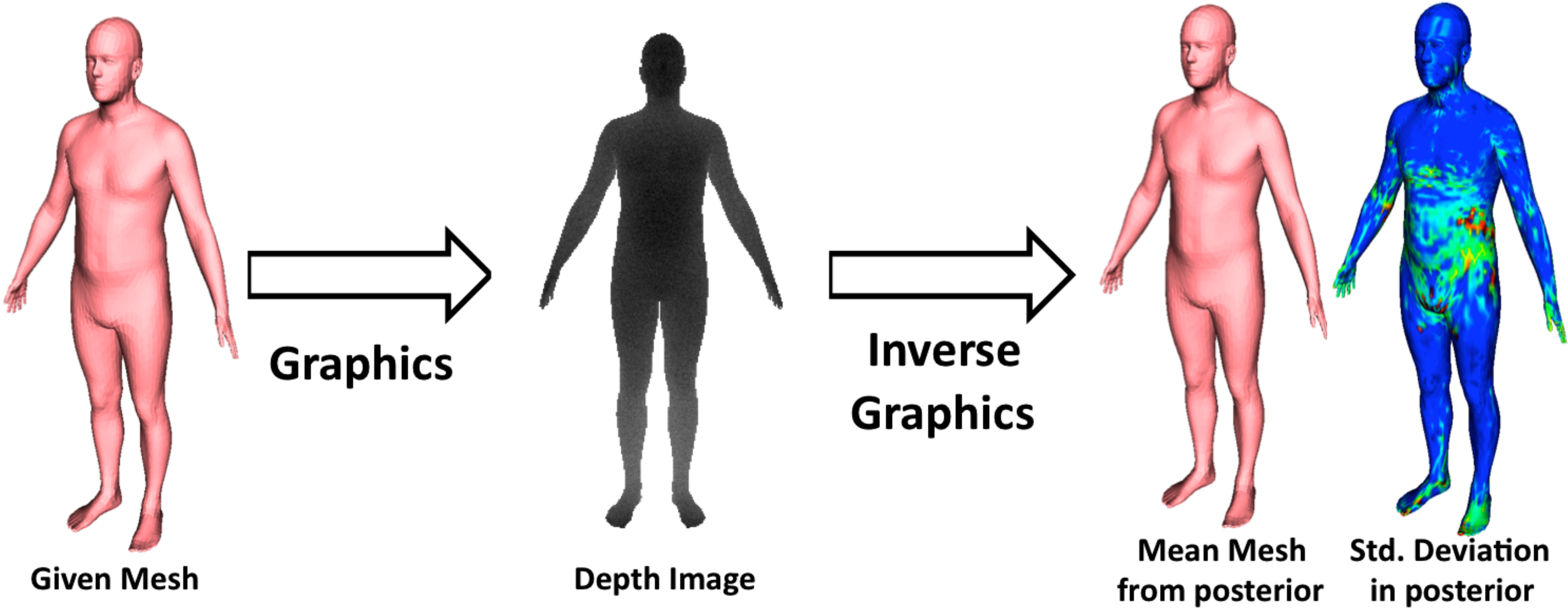}}
\caption{An example ``inverse graphics'' problem. A graphics engine renders a 3D
  body mesh and a depth image using an artificial camera. By Inverse
  Graphics we refer to the process of estimating the posterior
  probability over possible bodies given the depth image.}
\label{fig:teaser}
\end{center}
\end{figure}

Why did generative models not succeed? There are two key problems that
need to be addressed, the design of an accurate generative model, and
the inference therein. Modern computer graphic systems that leverage
dedicated hardware setups produce a stunning level of realism with
high frame rates. We believe that these systems will find its way in
the design of generative models and will open up exciting modelling
opportunities.
This observation motivates the research question of this paper, the
design of a general inference technique for efficient posterior
inference in accurate computer graphics systems. As such it can be
understood as an instance of \emph{Inverse
  Graphics}~\cite{baumgart1974inversegraphics}, illustrated in
Figure~\ref{fig:teaser} with one of our applications. 

The key problem in the generative world view is the difficulty of
posterior inference at test-time.
This difficulty stems from a number of reasons:
\emph{first}, the parameter $\theta$ is typically high-dimensional and so is
the posterior.
\emph{Second}, given $\theta$, the image formation process realizes complex and
\emph{dynamic} dependency structures, for example when objects occlude or
self-occlude each other.  These intrinsic ambiguities result in multi-modal
posterior distributions.
\emph{Third}, while most renderers are real-time, each simulation of the
forward process is expensive and prevents exhaustive enumeration.

We believe in the usefulness of generative models for computer vision
tasks, but argue that in order to overcome the substantial inference
challenges we have to devise techniques that are general and allow
reuse in several different models and novel scenarios. On the other
hand we want to maintain correctness in terms of the probabilistic
estimates that they produce. One way to improve on inference
efficiency is to leverage existing computer vision features and
discriminative models in order to aid inference in the generative
model.
In this paper, we propose the \emph{informed sampler}, a Markov chain
Monte Carlo (MCMC) method with discriminative proposal distributions.
It can be understood as an instance of a data-driven MCMC
method~\cite{zhu2000integrating}, and our aim is to design a method that
is general enough such that it can be applied across different
problems and is not tailored to a particular application.

During sampling, the informed sampler leverages computer vision features and
algorithms to make informed proposals for the state of latent variables and
these proposals are accepted or rejected based on the generative model.
The informed sampler is simple and easy to implement, but it enables inference
in generative models that were out of reach for current \emph{uninformed}
samplers.
We demonstrate this claim on challenging models that incorporate
rendering engines, object occlusion, ill-posedness, and
multi-modality. We carefully assess convergence statistics for the
samplers to investigate their truthfulness about the probabilistic
estimates. 
In our experiments we use existing computer vision technology:
our informed sampler uses standard histogram-of-gradients
features (HoG)~\citep{dalal2005histograms}, and the OpenCV
library, \citep{bradski2008opencv}, to produce informed proposals.
Likewise one of our models is an existing computer vision model,
the \emph{BlendSCAPE} model, a parametric model of human
bodies~\cite{hirshberg2012coregistration}.
In Section~\ref{sec:related}, we discuss related work and explain our informed
sampler approach in Section~\ref{sec:model}. Section~\ref{sec:experiments} presents baseline methods
and experimental setup. Then we present experimental analysis of informed sampler with three diverse
problems of estimating camera extrinsics (Section~\ref{sec:room}), occlusion reasoning (Section~\ref{sec:tiles}) and estimating body shape (Section~\ref{sec:bodyshape}). 
We conclude with a discussion of future work in Section~\ref{sec:discussion}.

\section{Related Work}
\label{sec:related}

This work stands at the intersection of computer vision, computer
graphics, and machine learning; it builds on previous approaches we
will discuss below. 

There is a vast literature on approaches to solve computer vision
applications by means of generative models. We mention some works that
also use an accurate graphics process as generative model. This
includes applications such as indoor scene
understanding~\cite{del2012bayesian}, human pose
estimation~\cite{lee2004proposal}, hand pose
estimation~\cite{de2008model} and many more. Most of these works are
however interested in inferring MAP solutions, rather than the full
posterior distribution.

Our method is similar in spirit to a \emph{Data Driven
  Markov Chain Monte Carlo} (DDMCMC) methods that use a bottom-up
approach to help convergence of MCMC sampling. DDMCMC methods have
been used in image segmentation~\cite{tu2002image}, object
recognition~\cite{zhu2000integrating}, and human pose
estimation~\cite{lee2004proposal}. The idea of making Markov samplers
data dependent is very general, but in the works mentioned above,
lead to highly problem specific implementations, mostly using
approximate likelihood functions. It is due to specialization on a
problem domain, that the proposed samplers are not easily transferable
to new problems. This is what we focus on in our work: to provide a simple,
yet efficient and general inference technique for problems where an accurate
forward process exists.
Because our method is general we believe that it is easy to adapt to a variety
of new models and tasks.

The idea to invert graphics~\cite{baumgart1974inversegraphics} in order to
understand scenes also has roots in the computer graphics community under the term
``inverse rendering''. The goal of inverse rendering however is to
derive a direct mathematical model for the forward light transport
process and then to analytically invert it. The work
of~\cite{ramamoorthi2001signal} falls in this category. The authors
formulate the light reflection problem as a convolution, to then
understand the inverse light transport problem as a deconvolution.
While this is a very elegant way to pose the problem, it does require
a specification of the inverse process, a requirement generative
modelling approaches try to circumvent.

Our approach can also be viewed as an instance of a probabilistic
programming approach. In the recent work
of~\cite{mansinghka2013approximate}, the authors combine graphics
modules in a probabilistic programming language to formulate an
approximate Bayesian computation. Inference is then implemented using
Metropolis-Hastings (\MH) sampling. This approach is appealing in its
generality and elegance, however we show that for our graphics
problems, a plain \MH sampling approach is not sufficient to achieve reliable
inference and that our proposed informed sampler can achieve robust
convergence in these challenging models.
Another piece of work from~\cite{stuhlmueller2013nips} is similar to our
proposed inference method in that knowledge about the forward process is
learned as ``stochastic inverses'', then applied for MCMC sampling in a
Bayesian network.
In the present work, we devise an MCMC sampler that we show works in both a
multi-modal problem as well as for inverting an existing piece of image
rendering code. In summary, our method can be understood in a similar
context as the above-mentioned papers,
including~\cite{mansinghka2013approximate}.


\section{The Informed Sampler}
\label{sec:model}

In general, inference about the posterior distribution is challenging
because for a complex model $p({\hat I}|\theta)$ no closed-form
simplifications can be made. This is especially true in the case that
we consider, where $p({\hat I}|\theta)$ corresponds to a graphics
engine rendering images. 
Despite this apparent complexity we observe the following: for many
computer vision applications there exist well performing
discriminative approaches, that, given the image, predict some
parameters $\theta$ or distributions thereof. These do not correspond
to the posterior distribution that we are interested in, but,
\emph{intuitively} the availability of discriminative inference
methods should make the task of inferring $p(\theta|{\hat I})$ easier.
\emph{Furthermore} a physically accurate generative model can be used in an
offline stage prior to inference to generate as many samples as we would like
or can afford computationally.  Again, \emph{intuitively} this should allow us
to prepare and summarize useful information about the distribution in order to
accelerate test-time inference.

Concretely, in our case we will use a discriminative method to provide a
global density $T_G({\hat I})$, which we then use in a valid MCMC inference method.
In the remainder of the section we first review Metropolis-Hastings Markov
Chain Monte Carlo (MCMC) and then discuss our proposed \emph{informed
samplers}.

\subsection{Metropolis-Hastings MCMC}
The goal of any sampler is to realize \textit{independent and identically
distributed} samples from a given probability distribution.
MCMC sampling, due to~\citet{metropolis1953} is a
particular instance that generates a sequence of random variables by
simulating a Markov chain.
Sampling from a target distribution $\pi(\cdot)$ consists of repeating the
following two steps~\cite{liu2001montecarlo}:

\newcommand{\thetaProp}{\bar{\theta}}
\begin{enumerate}
\item Propose a transition using a \textit{proposal distribution $T$}
  and the current state $\theta_t$
\begin{equation*}
\thetaProp \sim T(\cdot|\theta_t)
\end{equation*}
\item Accept or reject the transition based on Metropolis Hastings (\MH) acceptance rule: 
\begin{equation*}
\theta_{t+1} = \left\{
  \begin{array}{cl}
      \thetaProp,  & \textrm{rand}(0,1) < \min\left(1,\frac{\pi(\bar{\theta})
T(\thetaProp\to \theta_t)}{\pi(\theta_t) T(\theta_t \to \thetaProp)} \right), \\
     \theta_t, & \textrm{otherwise.}
  \end{array}
\right.
\end{equation*}
\end{enumerate}

Different MCMC techniques mainly differ in the implementation of the
proposal distribution $T$.

\subsection{Informed Proposal Distribution}
We use a common mixture kernel for Metropolis-Hastings sampling
\begin{equation}
T_{\alpha}(\cdot | {\hat I}, \theta_t) =
	\alpha \: T_L(\cdot | \theta_t) + (1-\alpha) \: T_G(\cdot |
        {\hat I}).
\label{eq:proposal}
\end{equation}
Here $T_L$ is an ordinary \emph{local} proposal distribution, for example a
multivariate Normal distribution centered around the current sample $\theta$,
and $T_G$ is a \emph{global} proposal distribution independent
of the current state. 
We inject knowledge by conditioning the global proposal
distribution $T_G$ on the image observation.
We learn the informed proposal $T_G(\cdot | \hat I)$ discriminatively in an
offline training stage using a non-parametric density estimator described
below.

The mixture parameter $\alpha \in [0,1]$ controls the contribution of each
proposal, for $\alpha=1$ we recover \MH.
For $\alpha=0$ the proposal $T_{\alpha}$ would be identical to $T_G(\cdot |
\hat I)$ and the resulting Metropolis sampler would be a valid metropolized
independence sampler~\citep{liu2001montecarlo}.
With $\alpha=0$ we call this baseline method \emph{Informed Independent \MH} (\INDLMH).
For intermediate values, $\alpha\in(0,1)$, we combine local with global moves
in a valid Markov chain.
We call this method \emph{Informed Metropolis Hastings} (\MIXLMH).

\subsection{Discriminatively Learning $T_G$}
The key step in the construction of $T_G$ is to include discriminative
information about the sample ${\hat I}$.
Ideally we would hope to have $T_G$ propose global moves which improve mixing
and even allow mixing between multiple modes, whereas the local proposal $T_L$ is
responsible for exploring the density locally.
To see that this is in principle possible, consider the case of a perfect
global proposal, that is,
$T_G(\cdot | {\hat I})=p_{\theta}(\cdot | {\hat I})$.
In that case we would get independent samples with $\alpha=0$ because every
proposal is accepted.
In practice $T_G$ is only an approximation to $p_{\theta}(\cdot|{\hat I})$.
If the approximation is good enough then the mixture of local and global
proposals will have a high acceptance rate and explore the density rapidly.

In principle we can use any conditional density estimation technique
for learning a proposal $T_G$ from samples.
Typically high-dimensional density estimation is difficult and even
more so in the conditional case; however, in our case we do have the
true generating process available to provide example pairs $(\theta,I)$. 
Therefore we use a simple but scalable non-parametric density
estimation method based on clustering a feature representation of the
observed image, $v({\hat I}) \in \mathbb{R}^d$. For each cluster we
then estimate an unconditional density over $\theta$ using kernel
density estimation (KDE). We chose this simple setup since it can
easily be reused in many different scenarios, in the experiments we
solve diverse problems using the same method. This method yields a
valid transition kernel for which detailed balance holds.

In addition to the KDE estimate for the global transition kernel we
also experimented with a random forest approach that maps the
observations to transition kernels $T_G$. More details will be given
in Section~\ref{sec:bodyshape}.

    \begin{algorithm}[t]
      \caption{Learning a global proposal $T_G(\theta|I)$}
      \label{alg:training}
      \begin{algorithmic}
        \STATE 1. Simulate $\{(\theta^{(i)},I^{(i)})\}_{i=1,\dots,n}$
        from $p(I|\theta) \: p(\theta)$ \STATE 2. Compute a feature
        representation $v(I^{(i)})$ \STATE 3. Perform k-means
        clustering of $\{v(I^{(i)})\}_i$
        \STATE 4. For each cluster $C_j \subset \{1,\dots,n\}$, fit a kernel \\
        density estimate $\textrm{KDE}(C_j)$ to the vectors
        $\theta^{\{C_j\}}$
      \end{algorithmic}
    \end{algorithm}

    \begin{algorithm}[t]
      \caption{\MIXLMH}
      \label{alg:sampling}
      \begin{algorithmic}
        \STATE \textbf{Input:} observed image $\hat{I}$ \STATE $T_L$
        $\leftarrow$ Local proposal distribution (Gaussian) \STATE $c$
        $\leftarrow$ cluster for $v(\hat{I})$ \STATE $T_G$ $\leftarrow
        KDE(c)$ (as obtained by Alg.~\ref{alg:training}) \STATE $T =
        \alpha T_L + (1-\alpha) T_G$
        \STATE Initialize $\theta_1$
        \FOR{$t=1$ {\bfseries to} $N-1$}
        \STATE 1. Sample $\thetaProp \sim T(\cdot)$
        \STATE 2. $\gamma =
        \min\left(1,\frac{\pi(\thetaProp|\hat{I}) T(\thetaProp \to
          \theta_t)}{\pi(\theta_t|\hat{I}) T(\theta_t \to
          \thetaProp)} \right)$
        \IF{rand$(0,1) < \gamma$}
        \STATE $\theta_{t+1} = \thetaProp$
        \ELSE
        \STATE $\theta_{t+1} = \theta_t$
        \ENDIF
        \ENDFOR
      \end{algorithmic}
    \end{algorithm}
For the feature representation we leverage successful discriminative
features and heuristics developed in the computer vision community. Different
task specific feature representations can be used in order to provide
invariance to small changes in $\theta$ and to nuisance parameters.
The main inference method remains the same across problems.

We construct the KDE for each cluster and we use a relatively small kernel bandwidth in
order to accurately represent the high probability regions in the posterior.
This is similar in spirit to using only high probability regions as ``darts''
in the \emph{Darting Monte Carlo} sampling technique
of~\citet{sminchisescu2011generalized}.
We summarize the offline training in Algorithm~\ref{alg:training}.

At test time, this method has the advantage that given an image ${\hat
  I}$ we only need to identify the corresponding cluster once using
$v({\hat I})$ in order to sample efficiently from the kernel
density $T_G$.
We show the full procedure in Algorithm~\ref{alg:sampling}.

This method yields a transition kernel that is a mixture kernel of a
reversible symmetric Metropolis-Hastings kernel and a metropolized
independence sampler. The combined transition kernel $T$ is hence also
reversible. Because the measure of each kernel dominates the support
of the posterior, the kernel is ergodic and has the correct stationary
distribution~\cite{brooks2011mcmchandbook}.
This ensures correctness of the inference and in the experiments we
investigate the efficiency of the different methods in terms of convergence
statistics. 

\section{Setup and Baseline Methods}
\label{sec:experiments}

In the remainder of the paper we demonstrate the proposed method in three different
experimental setups.
For all experiments, we use four parallel chains initialized at different
random locations sampled from the prior.
The reported numbers are median statistics over
multiple test images except when noted otherwise.

\subsection{Baseline Methods}

\paragraph{Metropolis Hastings (\MH)}
Described above, corresponds to $\alpha=1$, we use a symmetric diagonal
Gaussian distribution, centered at $\theta_t$.
\paragraph{Metropolis Hastings within Gibbs (\MHWG)}
We use a Metropolis Hastings scheme in a Gibbs sampler, that is, we draw from
one-dimensional conditional distributions for proposing moves and the Markov
chain is updated along one dimension at a time.  We further use a blocked
variant of this \MHWG sampler,
where we update blocks of dimensions at a time, and denote it by \BMHWG.
\paragraph{Parallel Tempering (\PT)}
We use Parallel Tempering to address the problem of sampling from
multi-modal
distributions~\cite{geyer1991paralleltempering,swendsen1986replicamontecarlo}.
This technique
is also known as ``replica exchange MCMC
sampling''~\cite{hukushima1996exchange}.
We run different parallel chains at different temperatures $T$, sampling
$\pi(\cdot)^{\frac{1}{T}}$ and at each sampling step propose to exchange two
randomly chosen chains. In our experiments we run three chains at
temperature levels $T\in\{1,3,27\}$ that were found to be best working
out of all combinations in $\{1,3,9,27\}$ for all experiments
individually. The highest temperature levels corresponds to an almost
flat distribution. 
\paragraph{Regeneration Sampler (\REGAMH)}
We implemented a regenerative MCMC method
\cite{mykland1995regeneration} that performs
adaption~\cite{gilks1998adaptive} of the proposal distribution during
sampling.
We use the mixture kernel (Eq.~\ref{eq:proposal}) as proposal distribution and
adapt only the global part $T_G(\cdot|\hat{I})$. This is initialized as
the prior over $\theta$ and at times of regeneration we fit a KDE to
the already drawn samples.
For comparison we used the same mixture coefficient $\alpha$ as for \MIXLMH (more details of this technique in~\ref{appendix:regsampler}).

\subsection{MCMC Diagnostics}\label{sec:mcmcdiagnostics}
We use established methods for monitoring the convergence of our MCMC
method~\cite{kass1998roundtable,flegal2008mcmcdiagnostics}. In
particular, we report different diagnostics. We compare the different
samplers with respect to the number of iterations instead of time. The
forward graphics process significantly dominates the runtime and
therefore the iterations in our experiments correspond linearly to the
runtime.

\paragraph{Acceptance Rate (AR)}
The ratio of accepted samples to the total Markov chain length.  The higher
the acceptance rate, the fewer samples we need to approximate the
posterior.  Acceptance rate indicates how well the proposal distribution
approximates the true distribution \emph{locally}.
\paragraph{Potential Scale Reduction Factor (PSRF)}
The PSRF diagnostics~\cite{gelman1992psrf,brooks1998convergence} is
derived by comparing within-chain variances with between-chain
variances of sample statistics.  For this, it requires independent
runs of multiple chains (4 in our case) in parallel. Because our
sample $\theta$ is multi-dimensional, we estimate the PSRF for each
parameter dimension separately and take the maximum PSRF value as
final PSRF value. A value close to one indicates that all chains
characterize the same distribution. This does not imply convergence,
the chains may all collectively miss a mode. However, a PSRF value
much larger than one is a certain sign of lack of convergence of the
chain. PSRF also indicates how well the sampler visits different modes of 
a multi-modal distribution.

\paragraph{Root Mean Square Error (RMSE)}
During our experiments we have access to the input parameters
$\theta^*$ that generated the image. To assess whether the posterior
distribution covers the ``correct'' value we report the RMSE between
the posterior expectation $\mathbb{E}_{p(\cdot |\hat{I})}[G(\cdot)]$
and the value $G(\theta^*)$ of the generating input. Since there is
noise being added to the observation we do not have access to the
ground truth posterior expectation and therefore this measure is only
an indicator. Under convergence all samplers would agree on the same
correct value.

\subsection{Parameter Selection}

For each sampler we individually selected hyper-parameters that gave
the best PSRF value after $10k$ iterations. In case the PSRF does not
differ for multiple values, we chose the one with highest acceptance
rate.  We include a detailed analysis of the baseline samplers and parameter
selection in the supplementary material.

\section{Experiment: Estimating Camera Extrinsics}
\label{sec:room}

\begin{figure}[!h]
\begin{center}
\centerline{\includegraphics[width=.7\columnwidth]{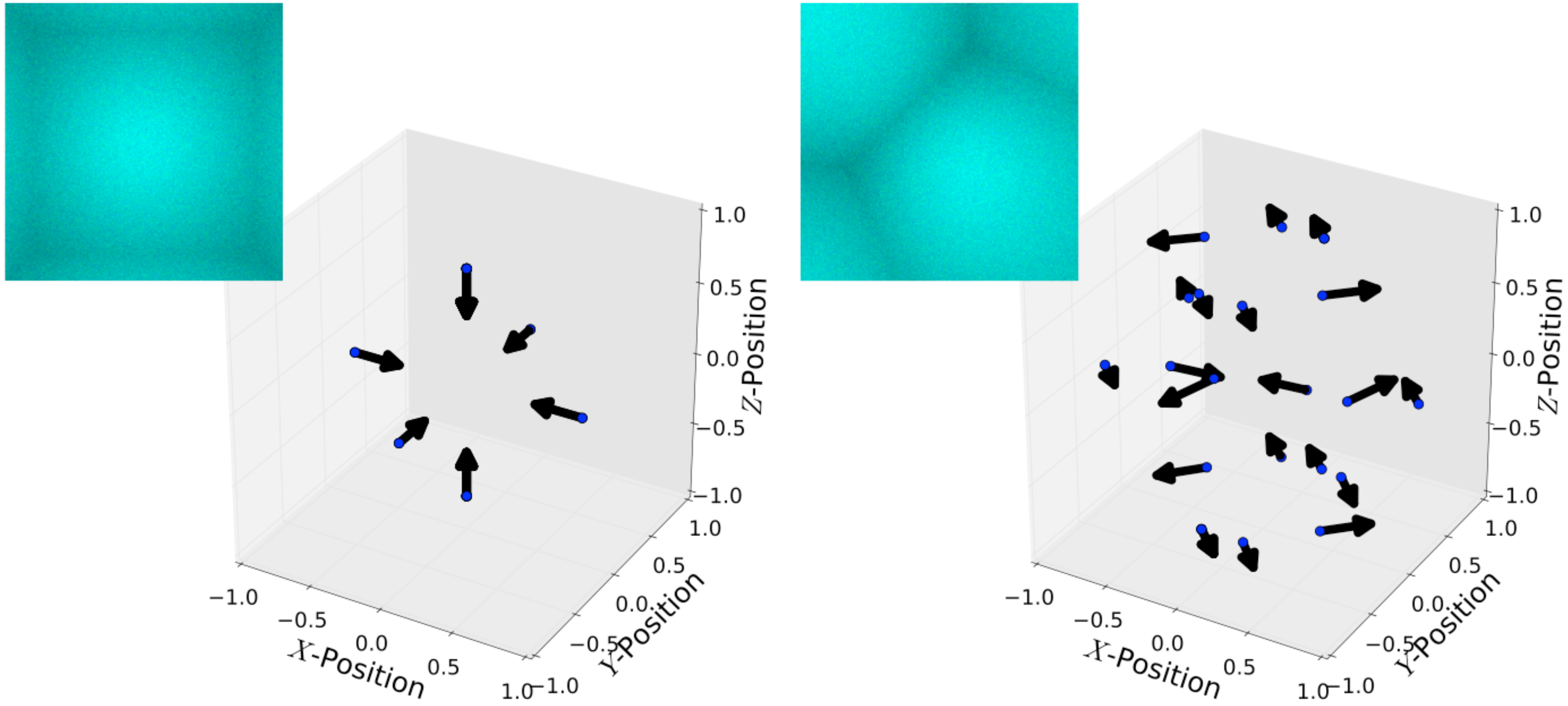}}
\vspace{-0.0cm}%
\caption{Two rendered room images with possible camera
  positions and headings that produce the same image. Not shown are
  the orientations; in the left example all six headings can
  be rolled by 90,180, and 270 degrees for the same image.\vspace{-.5cm}}
\label{fig:room}
\end{center}
\end{figure}

We implement the following simple graphics scenario to create a
challenging multi-modal problem. We render a cubical
room of edge length 2 with a point light source in the center of the
room $(0,0,0)$ from the viewpoint of a camera somewhere inside the
room. The camera parameters are described by its $(x,y,z)$-position
and the orientation, specified by yaw, pitch, and roll angles. The
inference process consists of estimating the posterior over these 6D
camera parameters $\theta$. See Figure~\ref{fig:room} for
two example renderings. Posterior inference is a highly multi-modal problem
because the room is a cubical and thus symmetric. There are 24
different camera parameters that will result in the same
image. This is also shown in Figure~\ref{fig:room} where we plot the
position and orientation (but not camera roll) of all camera
parameters that create the same image. A rendering of a
$200\times 200$ image with a resolution of $32bit$ using a single core
on an Intel Xeon 2.66GHz machine takes about $11ms$ on
average. 

A small amount of isotropic Gaussian noise is added to the rendered
image $G(\theta)$, using a standard deviation of $\sigma=0.02$. The
posterior distribution we try to infer then reads:
$p(\theta|\hat{I}) \propto p(\hat{I}|\theta)p(\theta) =
\mathcal{N}(\hat{I}|G(\theta),\sigma^2) \: \textrm{Uniform}(\theta)$.
The uniform prior over location parameters ranges between $-1.0$ and $1.0$ and 
the prior over angle parameters is modelled with wrapped uniform distribution over $[-\pi,\pi]$.

To learn the informed part of the proposal distribution from data, we
computed a histogram of oriented gradients (HOG)
descriptor~\cite{dalal2005histograms} from the image, using 9
gradient orientations and cells of size $20\times20$ yielding a feature vector
$v(I)\in\mathbb{R}^{900}$. We generated $300k$ training
images using a uniform prior over the camera extrinsic parameters, and
performed k-means using 5k cluster centers based on the HOG feature
vector. For each cluster cell, we then computed and stored a KDE for
the 6 dimensional camera parameters, following the steps in
Algorithm~\ref{alg:training}. 
As test data, we create 30 images using extrinsic parameters sampled
uniform at random over their range. 

\vspace{-0.2cm}
\subsection{Results}
\label{sec:roomresults}

\begin{figure*}[!t]
\begin{center}
\centerline{\includegraphics[width=\textwidth]{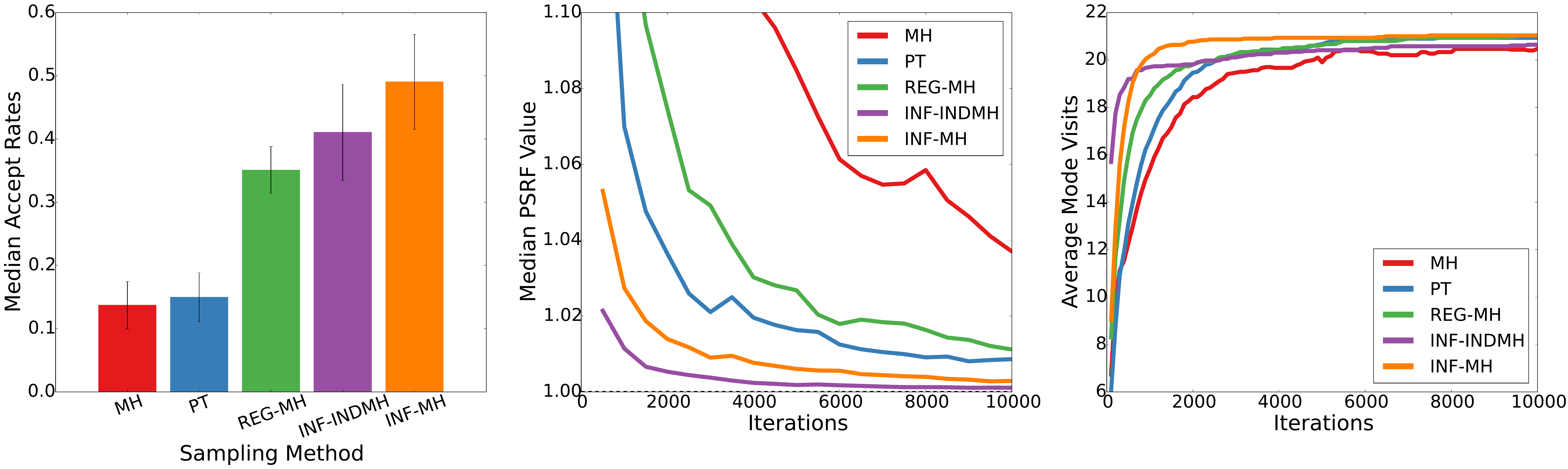}}
\caption{Results of the `Estimating Camera Extrinsics' experiment.
  Acceptance Rates (left), PSRFs (middle), and Average number of modes
  visited (right) for different sampling methods. We plot the
  median/average statistics over 30 test examples.\vspace{-.5cm}}
\label{fig:camPose_ALL}
\end{center}
\end{figure*}

We show results in Figure~\ref{fig:camPose_ALL}. We observe that both
\MH and \PT yield low acceptance rate compared to other methods.
However parallel tempering appears to overcome the multi-modality
better and improves over \MH in terms of convergence. The same holds
for the regeneration technique, we observe many regenerations, good
convergence and AR. Both \INDLMH and \MIXLMH converge quickly. 

In this experimental setup have access to the different exact modes,
there are 24 different ones. We analyze how quickly the samplers visit
the modes and whether or not they capture all of them. For ever
different instance the pairwise distances between the modes changes,
therefore we chose to define ``visiting a mode'' in the following way.
We compute a Voronoi tesselation with the modes as centers. A mode is
visited if a sample falls into its corresponding Voronoi cell, that
is, it is closer than to any other mode. Sampling uniform at random
would quickly find the modes (depending on the cell sizes) but is not
a valid sampler. We also experimented with balls of different radii
around the modes and found a similar behaviour to the one we report
here. Figure~\ref{fig:camPose_ALL} (right) shows results for various
samplers. We find that \MIXLMH discovers different modes quicker when
compared to other baseline samplers. Just sampling from the global
proposal distribution \INDLMH is initially visiting more modes (it is
not being held back by local steps) but is dominated by \MIXLMH over
some range. This indicates that the mixture kernel takes advantage of
both local and global moves, either one of them is exploring slower.
Also in most examples all samplers miss some modes under our
definition, the average number of discovered modes is 21 for \MIXLMH
and even lower for \MH. 

Figure~\ref{fig:exp1_alpha} shows the effect of mixture
coefficient ($\alpha$) on the informed sampling
\MIXLMH. Since there is no significant difference in PSRF values for
$0 \le \alpha \le 0.7$, we chose $0.7$ due to its high acceptance
rate.
Likewise, the parameters of the baseline samplers are chosen based on the PSRF and acceptance rate metrics.
See supplementary material for the analysis of the baseline samplers and the parameter selection.

\begin{figure}[h]
\centerline{\includegraphics[width=0.5\textwidth]{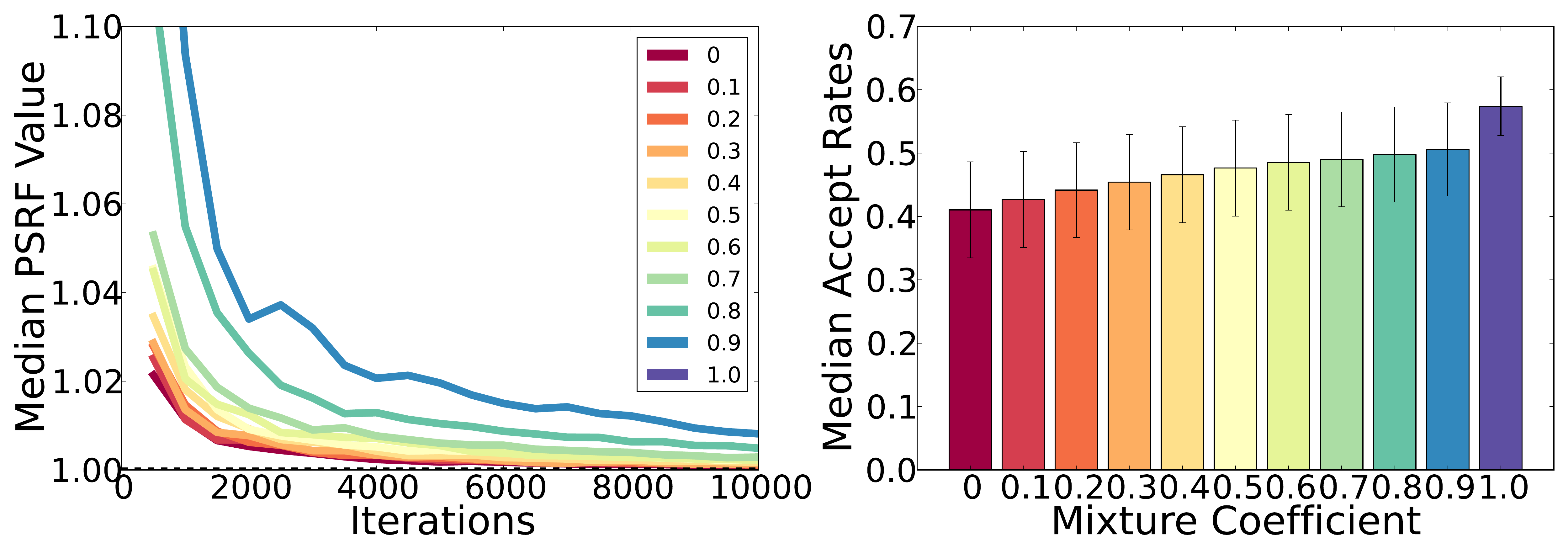}} 
\caption{Role of mixture coefficient. PRSFs and Acceptance rates corresponding to various mixture coefficients ($\alpha$) of \MIXLMH sampling in `Estimating Camera Extrinsics' experiment.} 
\label{fig:exp1_alpha}
\end{figure}

\begin{figure*}[!t]
\begin{center}
\centerline{\includegraphics[width=\textwidth]{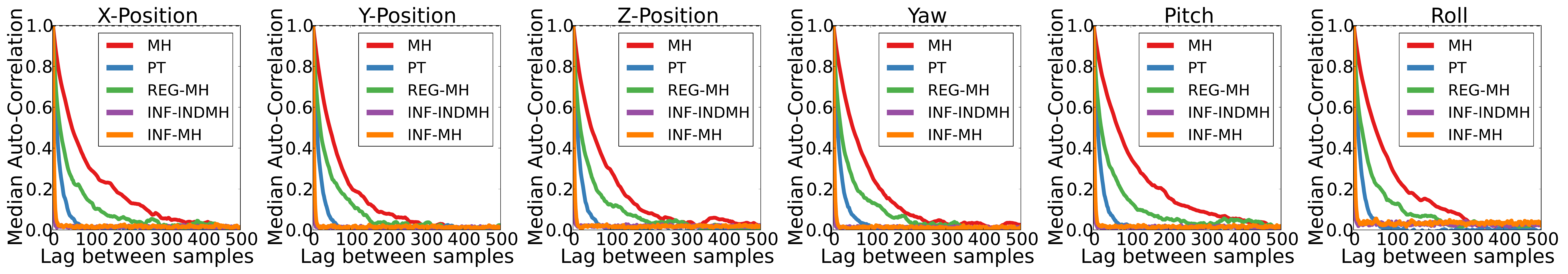}}
\caption{Auto-Correlation of samples obtained by different
      sampling techniques in camera extrinsics experiment, for each of
      the six extrinsic camera parameters.\vspace{-0.7cm}}
\label{fig:camPose_AC}
\end{center}
\end{figure*}

We also tested the \MHWG sampler and found that it did not converge
even after $100k$ iterations, with a PSRF value around 3. This is to be
expected since single variable updates will not traverse the
multi-modal posterior distributions fast enough due to the high
correlation of the camera parameters. In Figure~\ref{fig:camPose_AC} we plot the median
auto-correlation of samples obtained by different sampling techniques,
separately for each of the six extrinsic camera parameters. The
informed sampling approach (\MIXLMH~and \INDLMH) appears to produce samples which
are more independent compared to other baseline samplers.

As expected, some knowledge of the multi-modal structure of the
posterior needs to be available for the sampler to perform
well. The methods \INDLMH and \MIXLMH have this
information and perform better than baseline methods and \REGMH. 

\section{Experiment: Occluding Tiles}
\label{sec:tiles}

\begin{figure*}[th]
\begin{center}
\centerline{\includegraphics[width=\textwidth]{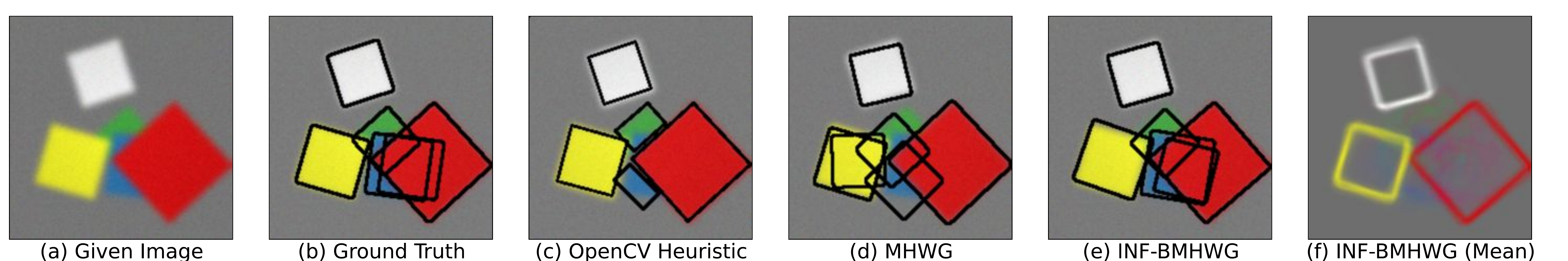}}
\caption{A visual result in `Occluding Tiles' experiment. (a) A sample rendered image, (b) Ground truth squares, and
  most probable estimates from 5000 samples obtained by (c) \MHWG
  sampler (best baseline) and (d) the \INFBMHWG sampler. (f)
  Posterior expectation of the square boundaries obtained by \INFBMHWG
  sampling. (The first 2000 samples are discarded as burn-in)\vspace{-.8cm}}
\label{fig:occlusion_main}
\end{center}
\end{figure*}

In a second experiment we render images depicting a fixed number
of six quadratic tiles placed at a random location $x$, $y$ in the image at a
random depth $z$ and orientation $\theta$. We blur the image and add
a bit of Gaussian random noise ($\sigma = 0.02$). An example is
depicted in Figure~\ref{fig:occlusion_main}(a), note that all the
tiles are of the same size, but farther away tiles look smaller. A rendering
of one $200 \times 200$ image takes about $25\textrm{ms}$ on average. Here, as prior,
we again use the uniform distribution over the 3D cube for tile location parameters, and
wrapped uniform distribution over $[-\frac{\pi}{4},\frac{\pi}{4}]$ for tile orientation angle.
To avoid label switching issues, each tile is given a fixed color and is not changed during the inference.

We chose this experiment such that it resembles the ``dead leaves model''
of~\citet{lee2001occlusion}, because it has properties that are
commonplace in computer vision. It is a scene composed of several
objects that are independent, except for occlusion, which complicates
the problem. If occlusion did not exist, the task is readily solved
using a standard OpenCV~\cite{bradski2008opencv} rectangle finding
algorithm ($\textrm{minAreaRect}$). The output of such an algorithm can be seen
in Figure~\ref{fig:occlusion_main}(c), and we use this algorithm as a discriminative
source of information. This problem is higher dimensional than the
previous one (24, due to 6 tiles of 4 parameters). Inference becomes
more challenging in higher dimension and our approach without
modification does not scale well with increasing dimensionality. One
way to approach this problem, is to factorize the joint distribution
into blocks and learn informed proposals separately. In the present
experiment, we observed that both baseline samplers and the plain
informed sampling fail when proposing all parameters jointly. Since
the tiles are independent except for the occlusion, we can approximate
the full joint distribution as product of block distributions where
each block corresponds to the parameters of a single tile. To estimate the full
posterior distribution, we learn global proposal distributions for
each block separately and use a block-Gibbs like scheme in our sampler
where we propose changes to one tile at a time, alternating between tiles.

The experimental protocol is the same as before, we render
500$k$ images, apply the OpenCV algorithm to fit rectangles and take their
found four parameters as features for clustering (10$k$ clusters). 
Again KDE distributions are fit to each cluster and at test time, we assign
the observed image to its corresponding
cluster.  The KDE in that chosen cluster determines the global sampler $T_G$
for that tile.  We then use $T_G$ to propose an update to all 4 parameters of
the tile. We refer to this procedure as
\INFBMHWG. Empirically we find $\alpha = 0.8$ to be optimal for \INFBMHWG sampling.

\subsection{Results}

An example result is shown in Figure~\ref{fig:occlusion_main}. We
found that the the \MH and \MIXLMH samplers fail entirely on this
problem. Both use a
proposal distribution for the entire state and due to the high
dimensions there is almost no acceptance ($< 1\%$) and thus they do
not reach convergence. The \MHWG sampler, updating one dimension at a time, is
found to be the best among the baseline samplers with acceptance rate of around $42\%$, 
followed by a block sampler that samples each tile separately.
The OpenCV algorithm produces a reasonable initial guess but fails in
occlusion cases.

\begin{figure*}[t]
\begin{center}
\centerline{\includegraphics[width=\textwidth]{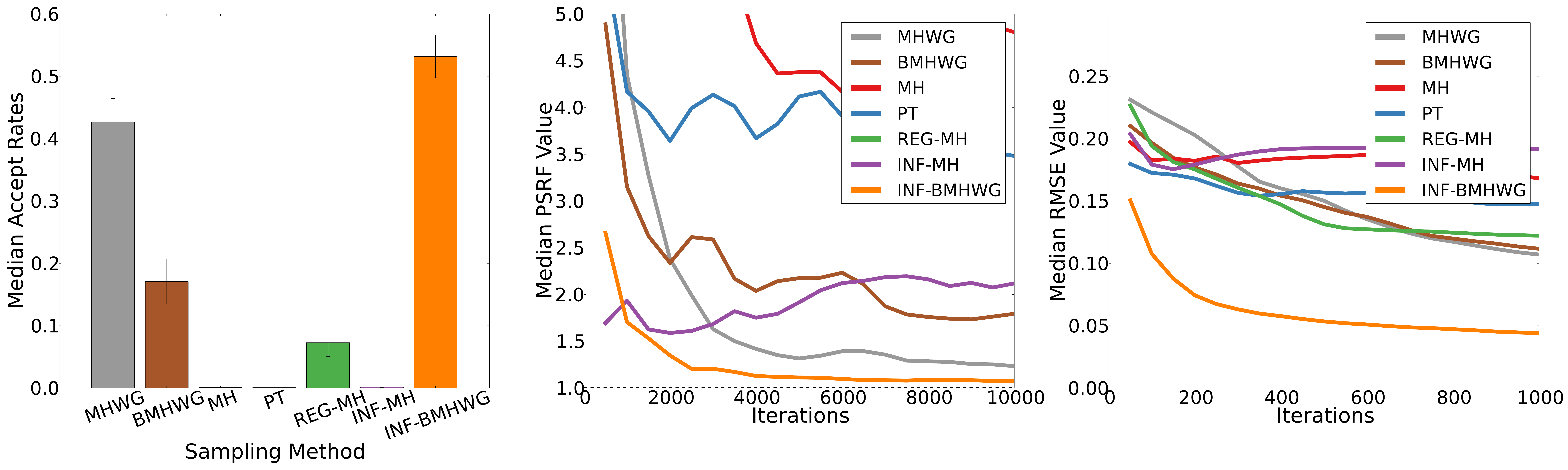}}
\caption{Results of the `Occluding Tiles' experiment. Acceptance Rates (left), PSRFs (middle), and RMSEs (right)
  for different sampling methods. Median results for 10 test examples.\vspace{-0.5cm}}
\label{fig:occlusion_ALL}
\end{center}
\end{figure*}

The block wise informed sampler \INFBMHWG converges quicker, with
higher acceptance rates ($\approx 53\%$), and lower reconstruction
error. The median curves for 10 test examples are shown in
Figure~\ref{fig:occlusion_ALL}, \INFBMHWG by far produces lower
reconstruction errors. Also in Fig~\ref{fig:occlusion_main}(f) the
posterior distribution is visualized, fully visible tiles are more
localized, position and orientation of occluded tiles more uncertain.
Figure~\ref{fig:exp2_visual} in the appendix shows some more visual results.
Although the model is relatively simple, all the baseline samplers
perform poorly and discriminative information is crucial to enable
accurate inference. Here the discriminative information is provided by a
readily available heuristic in the OpenCV library.

This experiment illustrates a variation of the informed sampling
strategy that can be applied to sampling from high-dimensional
distributions. Inference methods for general high-dimensional
distributions is an active area of research and intrinsically difficult.
The occluding tiles experiment is simple but illustrates this point,
namely that all non-block baseline samplers fail. Block sampling is a common
strategy in such scenarios and many computer vision problems have such
block-structure. Again the informed sampler improves in convergence
speed over the baseline method. Other techniques that produce better
fits to the conditional (block-)marginals should give faster
convergence.

\section{Experiment: Estimating Body Shape}
\label{sec:bodyshape}

The last experiment is motivated by a real world problem: estimating
the 3D body shape of a person from a single static depth image. With
the recent availability of cheap active depth sensors, the use of RGBD
data has become ubiquitous in computer
vision~\cite{shao2013rgbd,han2013computervisionkinect}.

\begin{figure*}[t]
\begin{center}
\centerline{\includegraphics[width=0.9\textwidth]{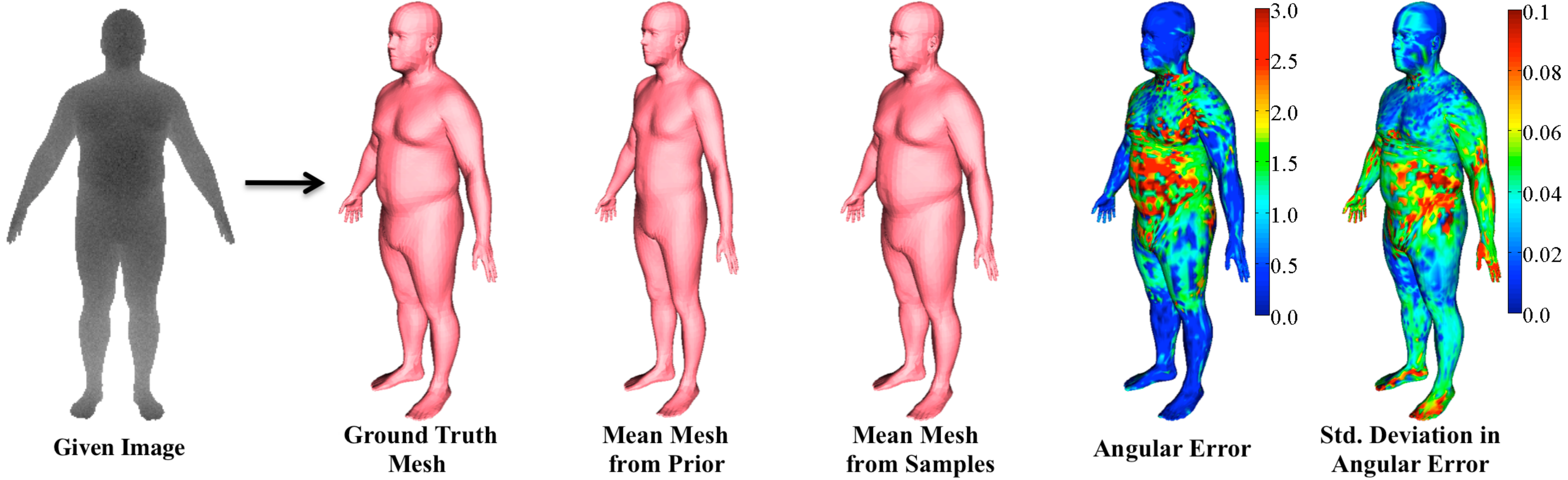}}
\caption{Inference of body shape from a depth image. A sample test result showing the result of 3D mesh
  reconstruction with the first 1000 samples obtained using our
  \MIXLMH sampling method. We visualize the angular error (in degrees) between the
  estimated and ground truth edge and project onto the mesh.}
\label{fig:meshVariances}
\end{center}
\end{figure*}

To represent a human body we use
the \emph{BlendSCAPE} model~\cite{hirshberg2012coregistration}, which updates the
originally proposed SCAPE model~\cite{anguelov2005scape} with better
training and blend weights. This model produces a 3D mesh of a human
body as shown in Figure~\ref{fig:meshVariances} as a function of
shape and pose parameters. The shape parameters allow us to represent
bodies of many builds and sizes, and includes a statistical
characterization (being roughly Gaussian). These parameters control
directions in deformation space, which were learned from a corpus of
roughly 2000 3D mesh models registered to scans of human bodies via
PCA. The pose parameters are joint angles which indirectly control
local orientations of predefined parts of the model.

Our model uses 57 pose parameters and any number of shape parameters
to produce a 3D mesh with 10,777 vertices. We use the first 7 SCAPE
components to represent the shape of a person. The camera
viewpoint, orientation, and pose of the person is held fixed. Thus a
rendering process takes $\theta\in\mathbb{R}^7$, generates a 3D mesh
representation of it and projects it through a virtual depth camera to
create a depth image of the person. This can be done in various
resolutions, we chose $430\times 260$ with depth values represented as
$32\textrm{bit}$ numbers in the interval $[0,4]$. On average, a full render path takes about
$28\textrm{ms}$. We add Gaussian noise with standard
deviation of $0.02$ to the created depth image. See
Fig.\ref{fig:meshVariances}(left) for an example.

We used very simple low level features for feature representation. In
order to learn the global proposal distribution we compute depth
histogram features on a $15\times 10$ grid on the image. For each cell
we record the mean and variance of the depth values. Additionally we
add the height and the width of the body silhouette as features
resulting in a feature vector $v(I)\in\mathbb{R}^{302}$. As
normalization, each feature dimension is divided by the maximum value
in the training set. We used $400k$ training images sampled from the
standard normal prior distribution and 10$k$ clusters to learn the KDE
proposal distributions in each cluster cell. 

For this experiment we also experimented with a different conditional
density estimation approach using a forest of random regression
trees~\cite{breiman1984cart,breiman2001randomforests}.
In the previous experiments, utilizing the KDE estimates, the
discriminative information entered through the feature representation.
Then, suppose if there was no relation between some observed features and the
variables that we are trying to infer, we would require a large number
of samples to reliably estimate the densities in the different
clusters. The regression forest can adaptively partition the parameter
space based on observed features and is able to ignore uninformative features, thus may lead to better
fits of the conditional densities. It can thus be understood as the
adaptive version of the k-Means clustering technique that solely
relies on the used metric (Euclidean in our case). 

In particular, we use the same features as for k-means clustering but
grow the regression trees using a mean square error criterion for
scoring the split functions. A forest of 10 binary trees with a depth
of 15 is grown, with the constraint of having a minimum of 40 training
points per leaf node. Then for each of the leaf nodes, a KDE is
trained as before. At test time the regression forest yields a mixture
of KDEs as the global proposal distribution. We denote this method as
\MIXLMHRF in the experiments.

Instead of placing using one KDE model for each cluster, we could also explore a regression approach, for example using a
discriminative linear regression model to map observations into
proposal distributions. By using informative covariates in the
regression model one should be able to overcome the curse of
dimensionality. Such a semi-parametric approach would allow to capture
explicit parametric dependencies of the variables (for example linear
dependencies) and combine them with non-parametric estimates of the
residuals. We are exploring this technique as future work.

Again, we chose parameters for all samplers individually, based on
empirical mixing rates. For informed samplers, we chose $\alpha = 0.8$, 
and a local proposal standard deviation of 0.05. The full analysis for all
samplers is included in the supplementary material. 

\vspace{-.2cm}
\subsection{Results}
\label{sec:bodyresults}

\begin{figure*}[t]
\begin{center}
\centerline{\includegraphics[width=\textwidth]{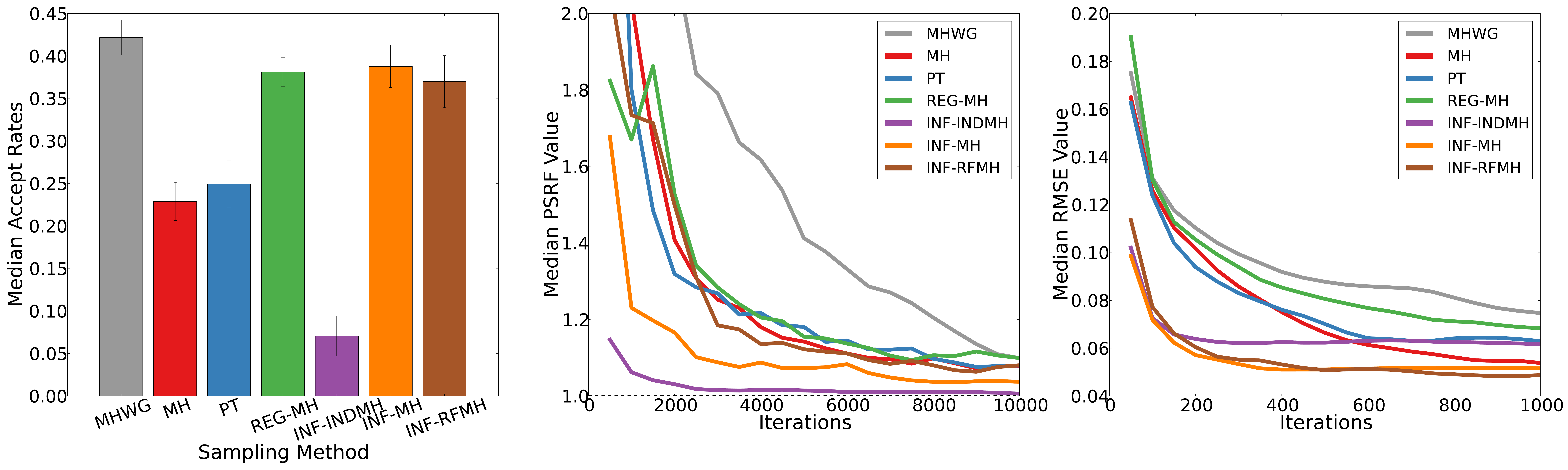}}
\caption{Results of the `Body Shape' experiment. Acceptance Rates (left), PSRFs (middle), and RMSEs (right)
  for different sampling methods in the body shape experiment. Median results over 10 test examples.\vspace{-0.5cm}}
\label{fig:bodyShape_ALL}
\end{center}
\end{figure*}

We tested the different approaches on 10 test images that are
generated by parameters drawn from the standard normal prior
distribution. Figure~\ref{fig:bodyShape_ALL} summarizes the results of the
sampling methods. We make the following observations. The baselines
methods \MH, \MHWG, and \PT show inferior convergence results and \MH and
\PT also suffer from lower acceptance rates. Just sampling from
the distribution of the discriminative step (\INDLMH) is not enough,
because the low acceptance rate indicates that the global proposals do not represent the
correct posterior distribution. However, combined with a local
proposal in a mixture kernel, we achieve a higher acceptance rate, faster convergence
and a decrease in RMSE. The regression forest approach has slower
convergence than \MIXLMH. In this example, the regeneration sampler
\REGMH does not improve over simpler baseline methods. We attribute
this to rare regenerations which may be improved with more specialized
methods.

We believe that our simple choice of depth image representation can
also significantly be improved on. For example, features can be
computed from identified body parts, something that the simple
histogram features have not taken into account. In the computer vision
literature some discriminative approaches for pose estimation do
exist, most prominent being the influential work on pose recovery in parts for
the Kinect XBox system~\cite{shotton2011kinect}.
In future work we plan to use similar methods to deal with pose variation and
complicated dependencies between parameters and observations.

\subsection{3D Mesh Reconstruction}

In Figure~\ref{fig:meshVariances} we show a sample 3D body mesh
reconstruction result using the \MIXLMH sampler after only 1000
iterations. We visualized the difference of the mean posterior and the
ground truth 3D mesh in terms of mesh edge directions. One can observe
that most differences are in the belly region and the feet of the
person. The retrieved posterior distribution allows us to assess the
model uncertainty.
To visualize the posterior variance we record standard deviation over
the edge directions for all mesh edges. This is backprojected to
achieve the visualization in Figure~\ref{fig:meshVariances}(right). We see
that posterior variance is higher in regions of higher error, that is, our
model predicts its own uncertainty correctly~\cite{dawid1982calibration}. In a
real-world body scanning scenario, this information will be
beneficial; for example, when scanning from multiple viewpoints or in
an experimental design scenario, it helps in selecting the next best
pose and viewpoint to record. Figure~\ref{fig:exp3_bodyMeshes} shows more 3D mesh reconstruction 
results using our sampling approach.

\subsection{Body Measurements}
\begin{figure}[!t]
  \begin{center}
    \centerline{\includegraphics[width=0.9\columnwidth]{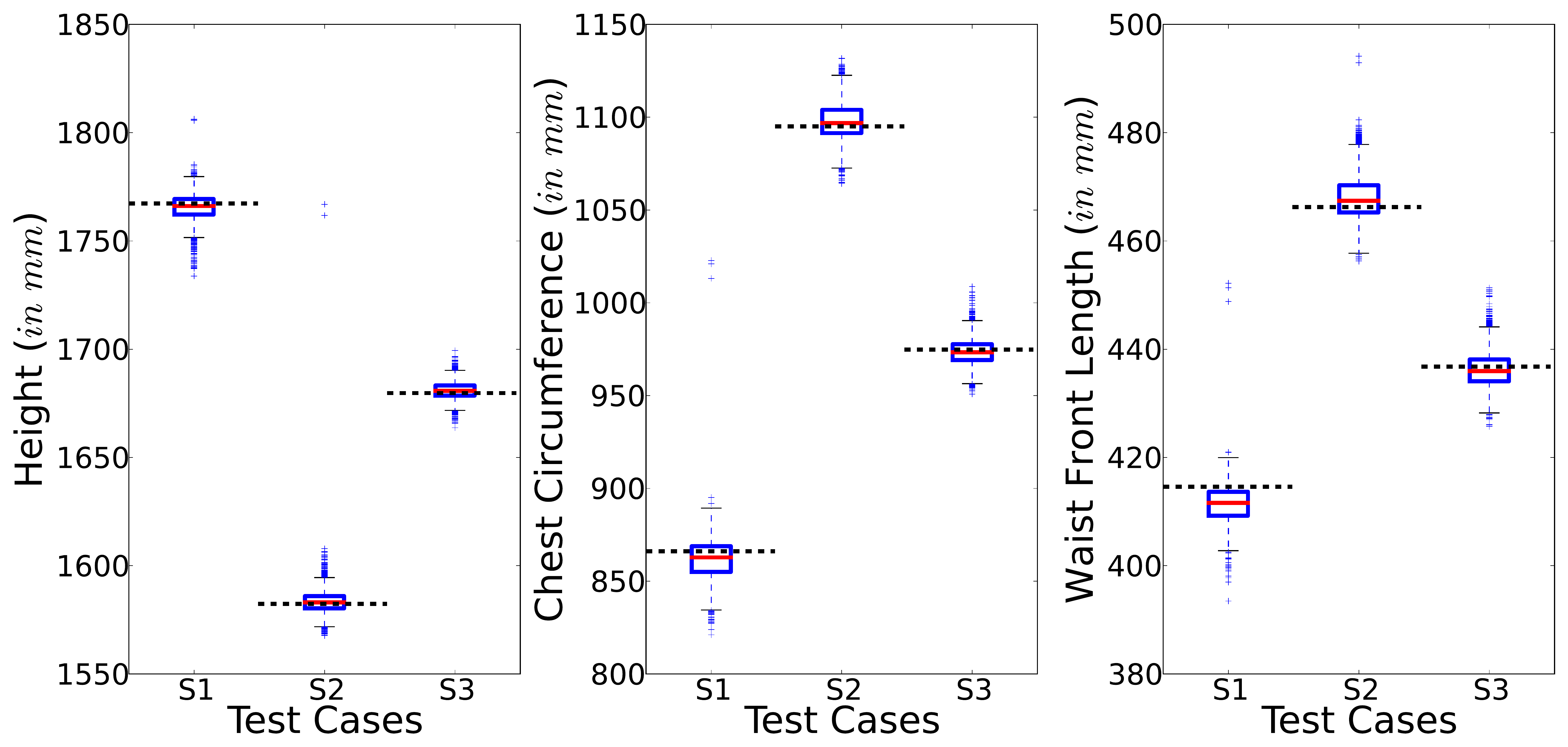}}
    \caption{Body measurements with quantified uncertainty. 
    Box plots of three body measurements for three test
      subjects, computed from the first $10k$ samples obtained by the
      \MIXLMH sampler. Dotted lines indicate measurements
      corresponding to ground truth SCAPE parameters.\vspace{-0.8cm}}
    \label{bodyMeasurements}
  \end{center}
\end{figure}

Predicting body measurements has many applications including clothing,
sizing and ergonomic design. Given pixel observations, one may wish to
infer a distribution over measurements (such as height and chest
circumference). Fortunately, our original shape training corpus
includes a host of 47 different per-subject measurements, obtained by
professional anthropometrists; this allows us to relate shape
parameters to measurements. Among many possible forms of regression,
regularized linear regression~\cite{zou2005regularization} was found
to best predict measurements from shape parameters. This linear
relationship allows us to transform any posterior distribution over
SCAPE parameters into a posterior over measurements, as shown in
Figure~\ref{bodyMeasurements}. We report for three randomly chosen
subjects' (S1, S2, and S3) results on three out of the 47
measurements. The dashed lines corresponds to ground truth
values. Our estimate not only faithfully recovers the true value but also
yields a characterization of the full conditional posterior.

\subsection{Incomplete Evidence}

\begin{figure}[!t]
\begin{center}
\centerline{\includegraphics[width=1.0\columnwidth]{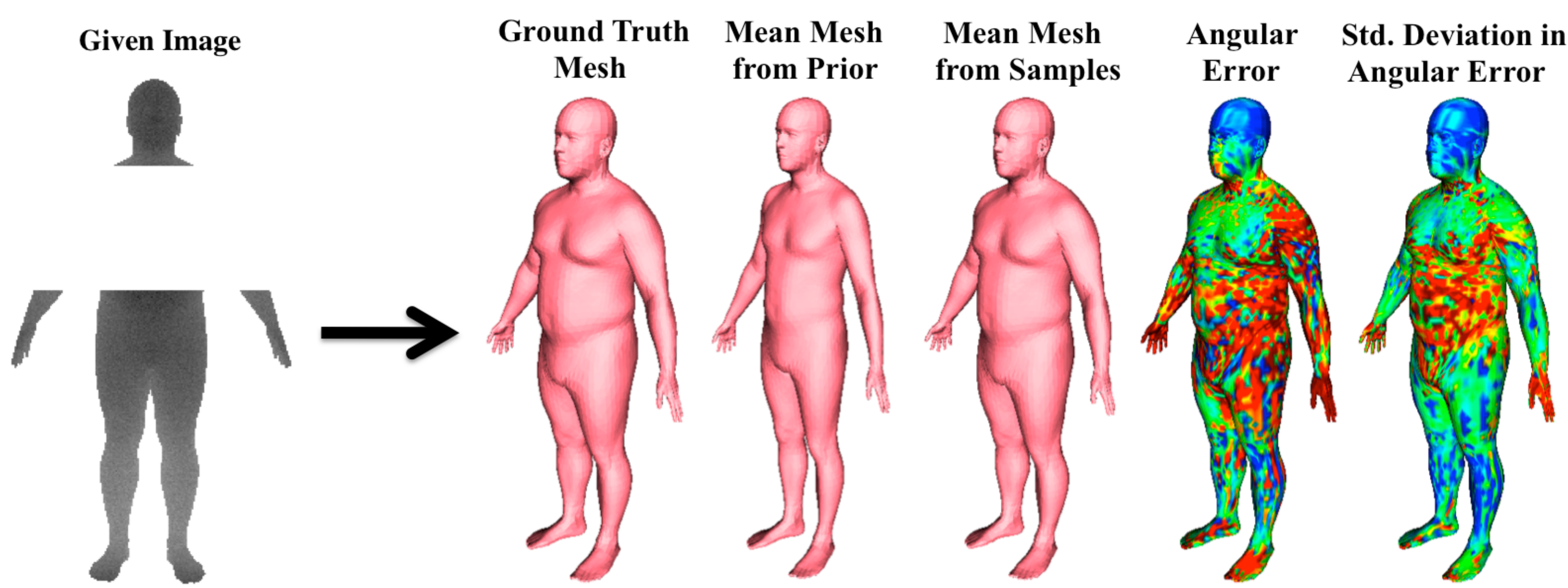}}
\caption{Inference with incomplete evidence. Mean 3D mesh and corresponding errors and uncertainties
  (std. deviations) in mesh edge directions, for the same test case as in figure~\ref{fig:meshVariances},
  computed from first $10k$ samples of our \MIXLMH sampling method
  with (bottom row) occlusion mask in image evidence. 
(blue indicates small values and red indicates high values)\vspace{-0.8cm}}
\label{fig:occlusionMeshes}
\end{center}
\end{figure}

Another advantage of using a generative model is the ability
to reason with missing observations. We perform a simple
experiment by occluding a portion of the observed depth image. 
We use the same inference and learning codes, with the same parametrization
and features as in the non-occlusion case but retrain the model to account for
the changes in the forward process. The result of \MIXLMH,
computed on the first $10k$ samples is shown in
Fig.~\ref{fig:occlusionMeshes}. The 3D reconstruction is reasonable even
under large occlusion; the error and the edge direction variance did
increase as expected.

\section{Discussion and Conclusions}
\label{sec:discussion}
This work proposes a method to incorporate discriminative methods into
Bayesian inference in a principled way. We augment a sampling
technique with discriminative information to enable inference with
global accurate generative models. Empirical results on three
challenging and diverse computer vision experiments are discussed. We
carefully analyse the convergence behaviour of several different
baselines and find that the informed sampler performs well across all
different scenarios. This sampler is applicable to general scenarios
and in this work we leverage the accurate forward process for offline
training, a setting frequently found in computer vision applications.
The main focus is the generality of the approach, this inference
technique should be applicable to many different problems and not be
tailored to a particular problem.

We show that even for very simple scenarios, most baseline samplers
perform poorly or fail completely. By including a global
image-conditioned proposal distribution that is informed through
discriminative inference we can improve sampling performance. We
deliberately use a simple learning technique (KDEs on k-Means cluster
cells and a forest of regression trees) to enable easy reuse in other
applications. Using stronger and more tailored discriminative models
should lead to better performance. We see this as a way where top-down
inference is combined with bottom-up proposals in a probabilistic
setting. 

There are some avenues for future work; we understand this method as
an initial step into the direction of general inference techniques for
accurate generative computer vision models. Identifying conditional
dependence structure should improve results, e.g.
recently~\citet{stuhlmueller2013nips} used structure in Bayesian
networks to identify such dependencies.
One assumption in our work is that we use an accurate generative
model. Relaxing this assumption to allow for more general scenarios
where the generative model is known only approximately is important
future work. In particular for high-level computer vision problems
such as scene or object understanding there are no accurate generative
models available yet but there is a clear trend towards physically
more accurate 3D representations of the world. 
This more general setting is different to the one we consider in this
paper, but we believe that some ideas can be carried over. For
example, we could create the informed proposal distributions from
manually annotated data that is readily available in many computer
vision data sets. Another problem domain are trans-dimensional models,
that require different sampling techniques like reversible jump MCMC
methods~\cite{green1995reversible,brooks2011mcmchandbook}. We are investigating general techniques to ``inform'' this
sampler in similar ways as described in this manuscript.

We believe that generative models are useful in many computer vision
scenarios and that the interplay between computer graphics and
computer vision is a prime candidate for studying probabilistic
inference and probabilistic
programming~\citep{mansinghka2013approximate}.
However, current inference techniques need to be improved on many
fronts: efficiency, ease of usability, and generality. Our method is a
step towards this direction: the informed sampler leverages the power
of existing discriminative and heuristic techniques to enable a
principled Bayesian treatment in rich generative models. Our emphasis
is on generality; we aimed to create a method that can be easily
reused in other scenarios with existing code bases.
The presented results are a successful example of the inversion of an
involved rendering pass.
In the future we plan to investigate ways to combine existing computer
vision techniques with principled generative models, with the aim of
being general rather than problem specific.

\section*{Appendix}
\appendix
\section{Regeneration Sampler (\REGMH)}
\label{appendix:regsampler}

Adapting the proposal distribution with existing MCMC samples is not
straight-forward as this would potentially violate the Markov property
of the chain~\cite{atchade2005adaptivemcmc}.
One approach is to identify \textit{times of
  regeneration} at which the chain can be restarted and the proposal
distribution can be adapted using samples drawn previously. Several approaches to identify good regeneration times in a
general Markov chain have been proposed~\cite{athreya1978new,
  nummelin1978splitting}. We build on~\cite{mykland1995regeneration}
that proposed two \textit{splitting} methods for finding the
regeneration times. Here, we briefly describe the method that we
implemented in this study. 

Let the present state of the sampler be $x$ and let the independent
global proposal distribution be $T_G$. When $y \sim T_G$ is accepted
according to the MH acceptance rule, the probability of a regeneration is
given by:
\begin{equation}
r(x,y) = \left\{
  \begin{array}{ll}
    \max\{ \frac{c}{w(x)}, \frac{c}{w(y)}  \},& \textrm{if $w(x)>c$ and $w(y)>c$},\\
    \max\{ \frac{w(x)}{c}, \frac{w(y)}{c}  \},& \textrm{if $w(x)<c$ and $w(y)<c$},\\
    1,    & \textrm{otherwise},
  \end{array}
\right.
\label{reg_eq}
\end{equation}
where $c > 0$ is an arbitrary constant and $w(x) =
\frac{\pi(x)}{T_G(x)}$. The value of $c$ can be set to maximize the
regeneration probability. At every sampling step, if a sample from the
independent proposal distribution is accepted, we compute regeneration
probability using equation~\ref{reg_eq}. If a regeneration occurs,
the present sample is discarded and replaced with one from the
independent proposal distribution $T_G$. We use the same mixture
proposal distribution as in our informed sampling approach where we
initialize the global proposal $T_G$ with a prior distribution and at
times of regeneration fit a KDE to the existing samples. This becomes
the new adapted distribution $T_G$. Refer to
\cite{mykland1995regeneration} for more details of this regeneration
technique. In the work of~\cite{ahn2013distributed} this regeneration
technique is used with success in a Darting Monte Carlo sampler.

\section{Additional Qualitative Results}

\subsection{Occluding Tiles}
\begin{figure*}[t]
\begin{center}
\centerline{\includegraphics[width=\textwidth]{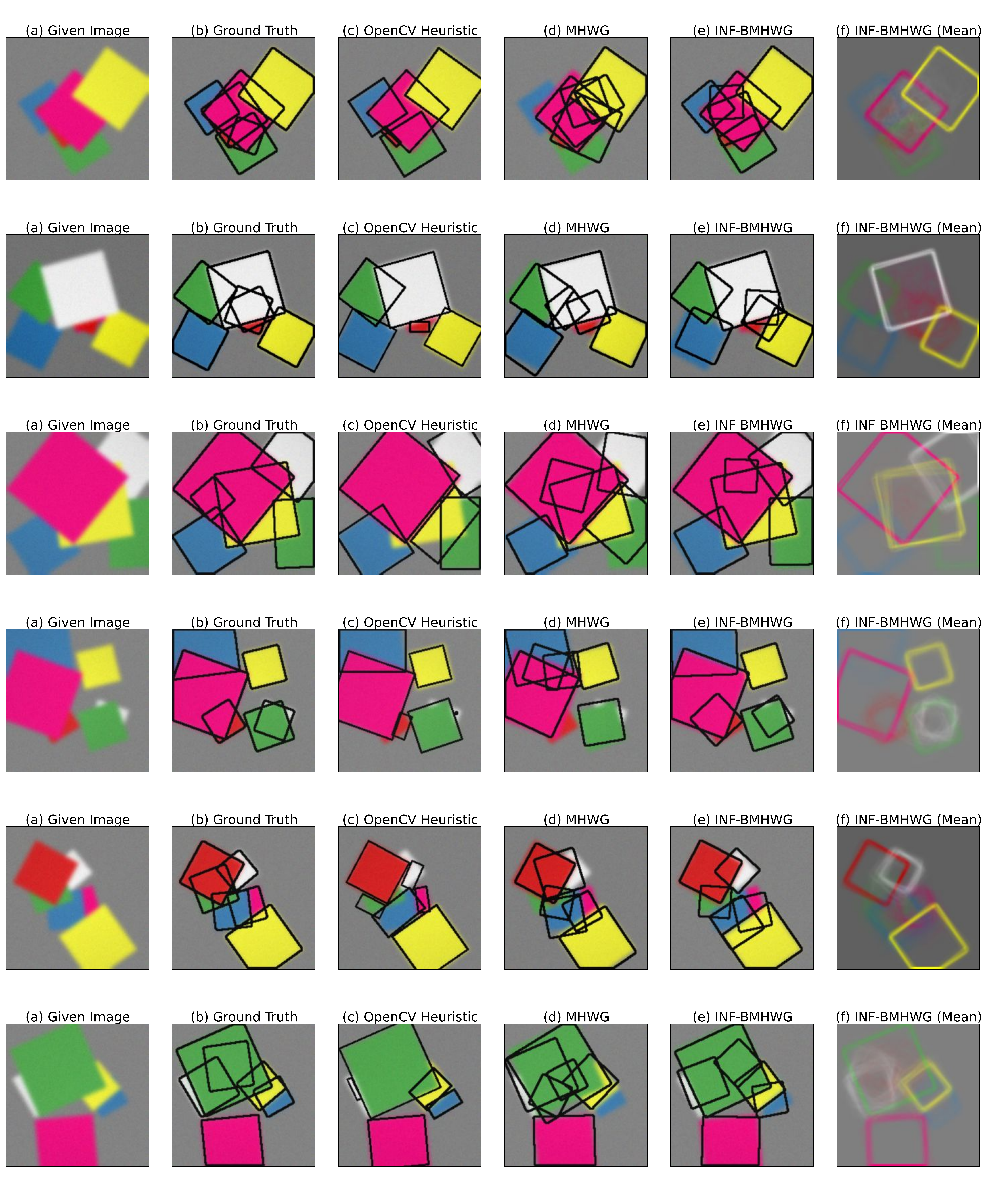}}
\caption{Qualitative Results of the occluding tiles experiment. From
  left to right: (a)
  Given image, (b) Ground truth tiles, and most probable estimates
  from 5000 samples obtained by (c) MHWG sampler (best baseline) and
  (d) our INF-BMHWG sampler. (f) Posterior expectation of the tiles
  boundaries obtained by INF-BMHWG sampling. (First 2000 samples are
  discarded as burn-in)}
\label{fig:exp2_visual}
\end{center}
\end{figure*}

In Figure~\ref{fig:exp2_visual} more qualitative results of the
occluding tiles experiment are shown. The informed sampling approach
(\INFBMHWG) is better than the best baseline (\MHWG). This still is a
very challenging problem since the parameters for occluded tiles are
flat over a large region. Some of the posterior variance of the
occluded tiles is already captured by the informed sampler.

\subsection{Body Shape}
\begin{figure*}[t]
\begin{center}
\centerline{\includegraphics[width=0.75\textwidth]{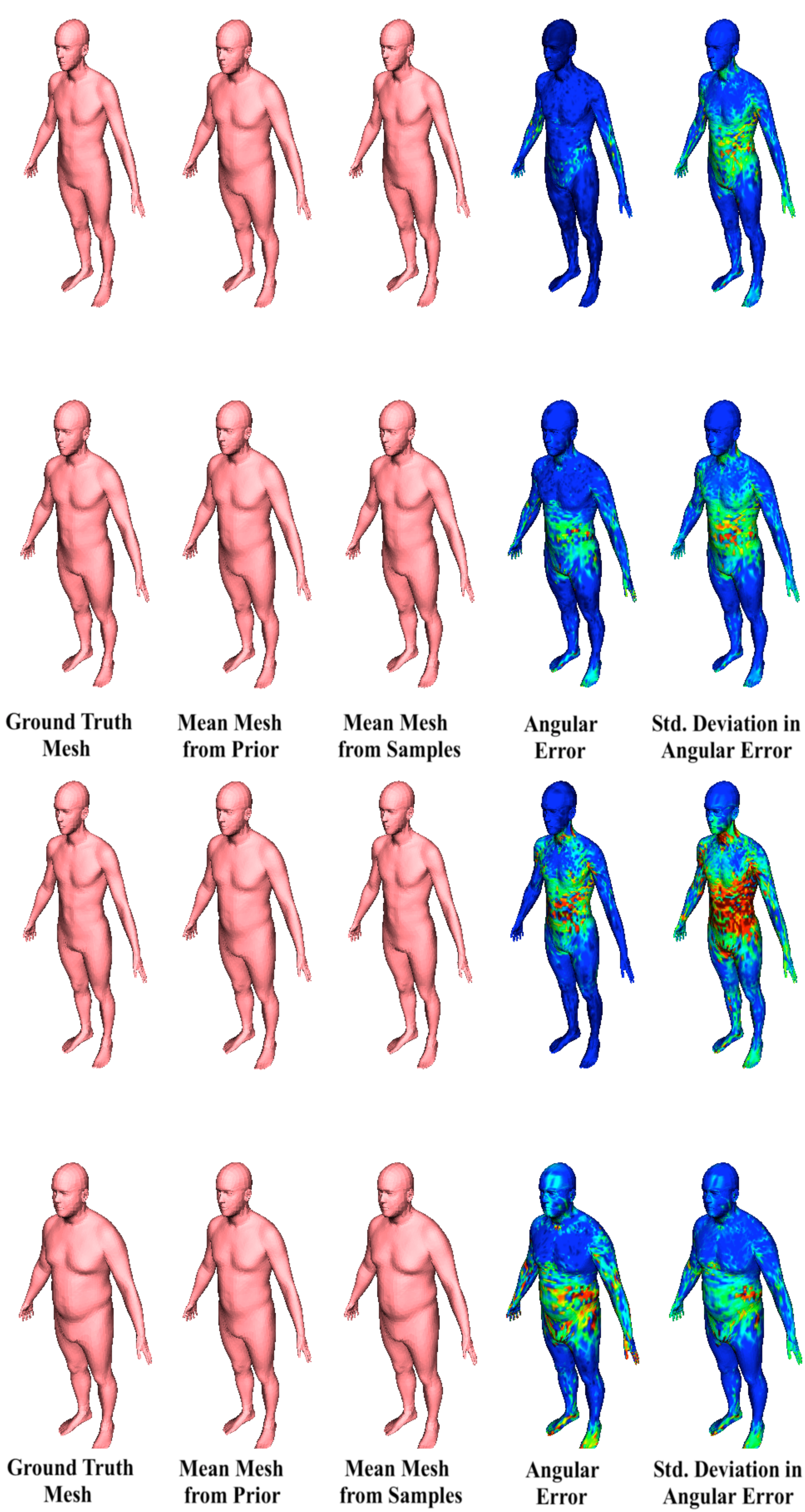}}
\caption{Qualitative results for the body shape experiment. Shown is
  the 3D mesh reconstruction results with first 1000 samples obtained
  using the \MIXLMH informed sampling method. (blue indicates small
  values and red indicates high values)}
\label{fig:exp3_bodyMeshes}
\end{center}
\end{figure*}

Figure~\ref{fig:exp3_bodyMeshes} shows some more results of 3D mesh
reconstruction using posterior samples obtained by our informed
sampling \MIXLMH.

\bibliographystyle{abbrvnat}
\bibliography{arxiv}

\newpage

\twocolumn[
\icmltitle{Supplementary Material for~\\ The Informed Sampler:~\\ A Discriminative Approach to Bayesian Inference in ~\\Generative Computer Vision Models}

\icmlauthor{Varun Jampani}{varun.jampani@tuebingen.mpg.de}
\icmladdress{Max Planck Institute for Intelligent Systems, Spemannstra{\ss}e 41, 72076, T{\"u}bingen, Germany}

\icmlauthor{Sebastian Nowozin}{sebastian.nowozin@microsoft.com}
\icmladdress{Microsoft Research Cambridge, 21 Station Road, Cambridge, CB1 2FB, United Kingdom}

\icmlauthor{Matthew Loper}{mloper@tuebingen.mpg.de}
\icmladdress{Max Planck Institute for Intelligent Systems, Spemannstra{\ss}e 41, 72076, T{\"u}bingen, Germany}

\icmlauthor{Peter V. Gehler}{pgehler@tuebingen.mpg.de}
\icmladdress{Max Planck Institute for Intelligent Systems, Spemannstra{\ss}e 41, 72076, T{\"u}bingen, Germany}

          \icmlkeywords{probabilistic models, markov chain monte
            carlo, inverse graphics, computer vision, generative
            models}

\vskip 0.3in
]

\section{Baseline Results and Analysis}

On the next pages of this supplementary material, we give an in-depth
performance analysis of the various samplers and the effect of their
hyperparameters. We choose hyperparameters with the lowest PSRF value
after $10k$ iterations, for each sampler individually. If the
differences between PSRF are not significantly different among
multiple values, we choose the one that has the highest acceptance
rate.



\section{Experiment: Estimating Camera Extrinsics}

\subsection{Parameter Selection}
\paragraph{Metropolis Hastings (\MH)}

\begin{figure}[h]
\centering
  \subfigure[MH]{%
    \includegraphics[width=.48\textwidth]{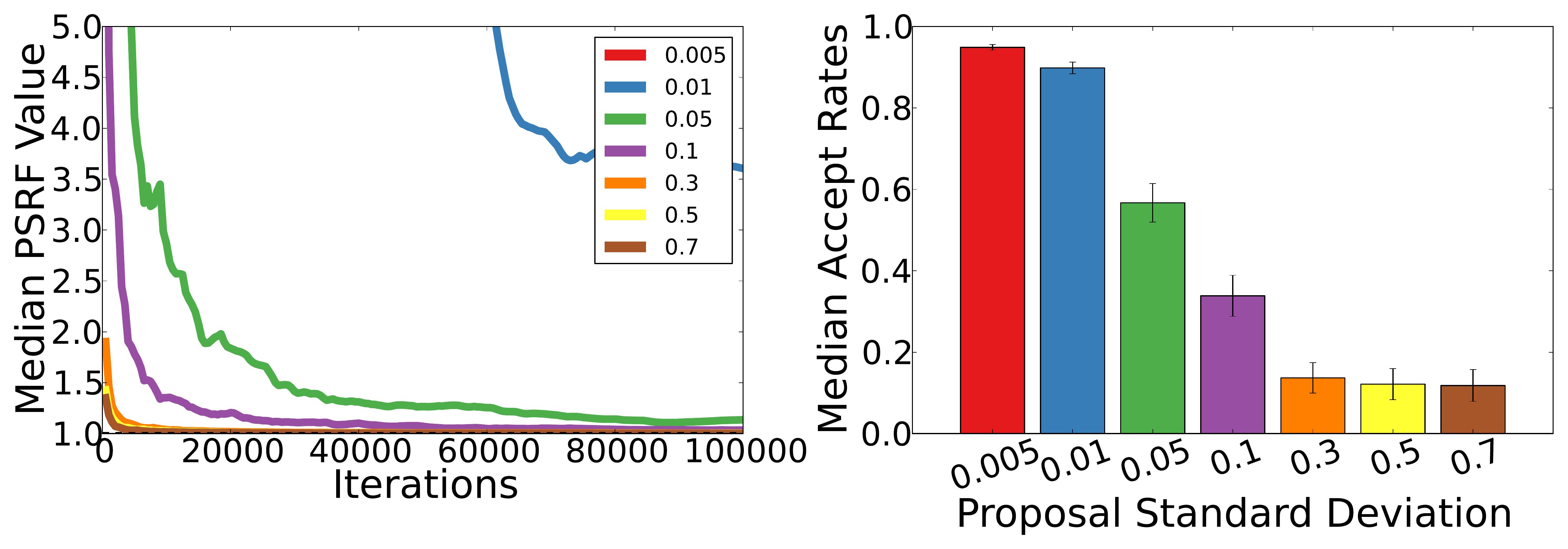} \label{fig:exp1_MH} 
  } 
  \subfigure[PT]{%
    \includegraphics[width=.48\textwidth]{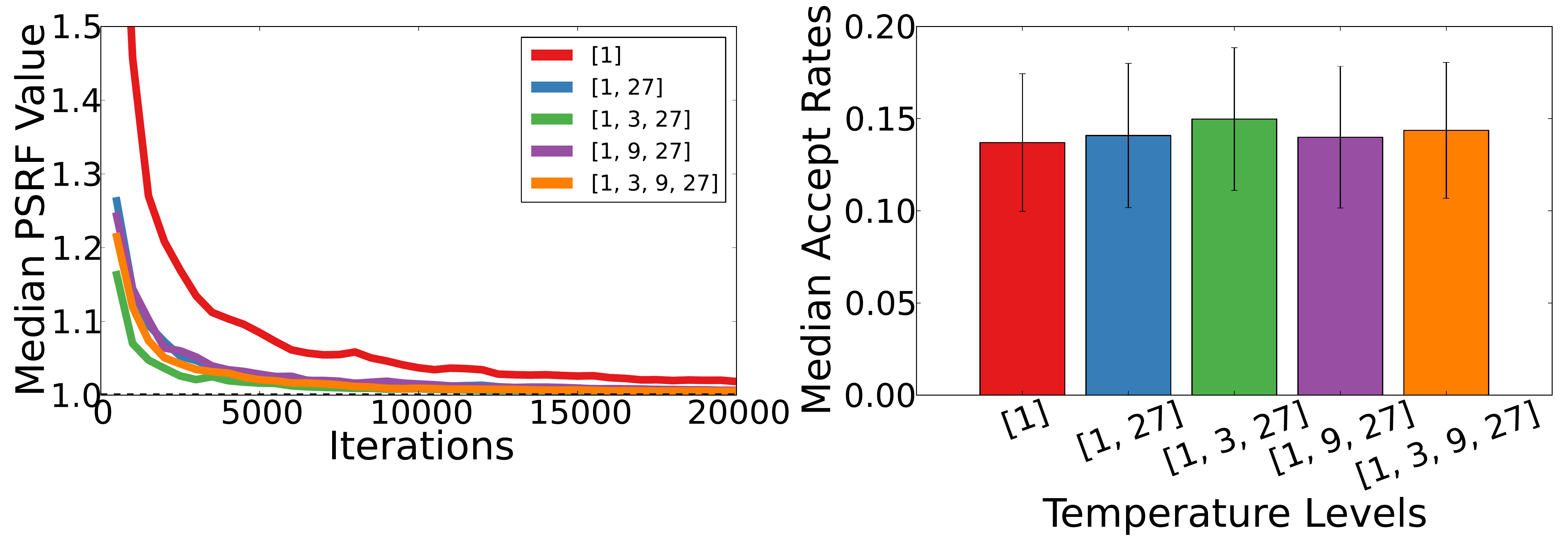} \label{fig:exp1_PT} 
  }
\\
  \subfigure[INF-MH]{%
    \includegraphics[width=.48\textwidth]{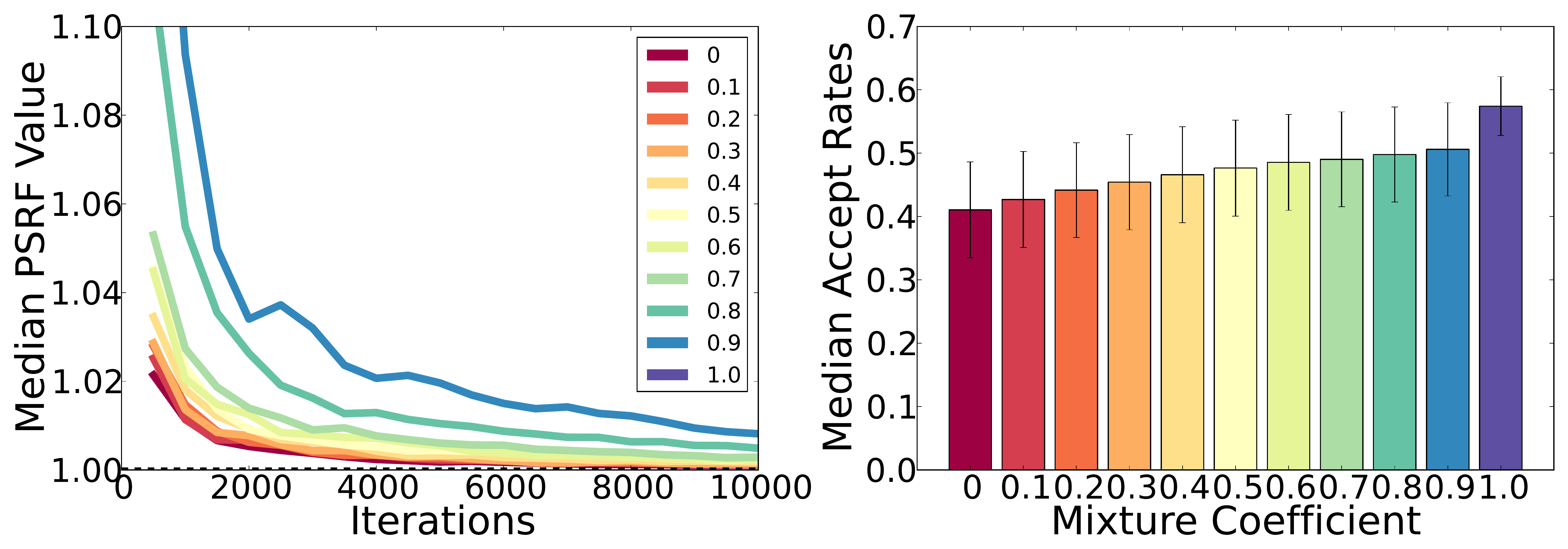} \label{fig:exp1_alpha} 
  }
  \caption{Results of the `Estimating Camera Extrinsics' experiment. PRSFs and Acceptance rates corresponding to (a) various standard deviations of \MH, (b) various temperature level combinations of \PT sampling and (c) various mixture coefficients of \MIXLMH sampling.} 
\end{figure}

Figure~\ref{fig:exp1_MH} shows the median acceptance rates and PSRF
values corresponding to various proposal standard deviations of plain
\MH~sampling. Mixing gets better and the acceptance rate gets worse as
the standard deviation increases. The value $0.3$ is selected standard
deviation for this sampler.

\paragraph{Metropolis Hastings Within Gibbs (\MHWG)}

As mentioned in the main paper, the \MHWG~sampler with one-dimensional
updates did not converge for any value of proposal standard deviation.
This problem has high correlation of the camera parameters and is of
multi-modal nature, which this sampler has problems with. 

\paragraph{Parallel Tempering (\PT)}

For \PT~sampling, we took the best performing \MH~sampler and used
different temperature chains to improve the mixing of the
sampler. Figure~\ref{fig:exp1_PT} shows the results corresponding to
different combination of temperature levels. The sampler with
temperature levels of $[1,3,27]$ performed best in terms of both
mixing and acceptance rate.

\paragraph{Effect of Mixture Coefficient in Informed Sampling (\MIXLMH)}

Figure~\ref{fig:exp1_alpha} shows the effect of mixture
coefficient ($\alpha$) on the informed sampling
\MIXLMH. Since there is no significant different in PSRF values for
$0 \le \alpha \le 0.7$, we chose $0.7$ due to its high acceptance
rate.


\begin{figure}[!t]
\centering
  \subfigure[\MH]{%
    \includegraphics[width=.48\textwidth]{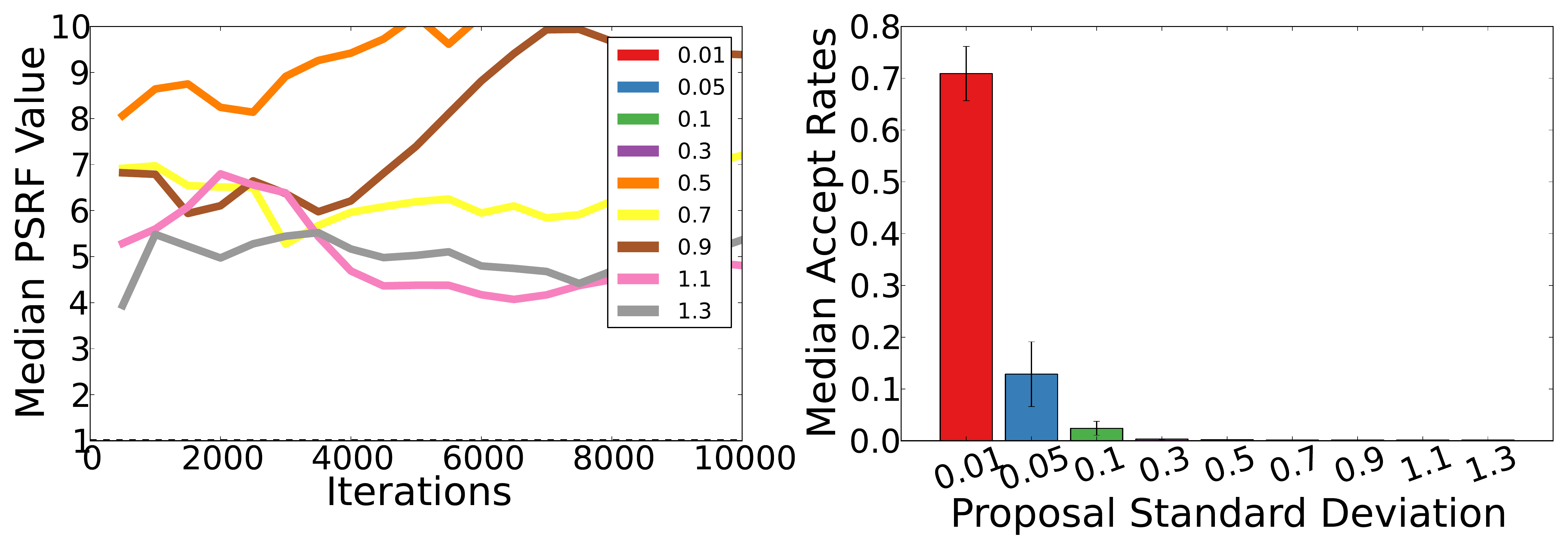} \label{fig:exp2_MH} 
  } 
  \subfigure[\BMHWG]{%
    \includegraphics[width=.48\textwidth]{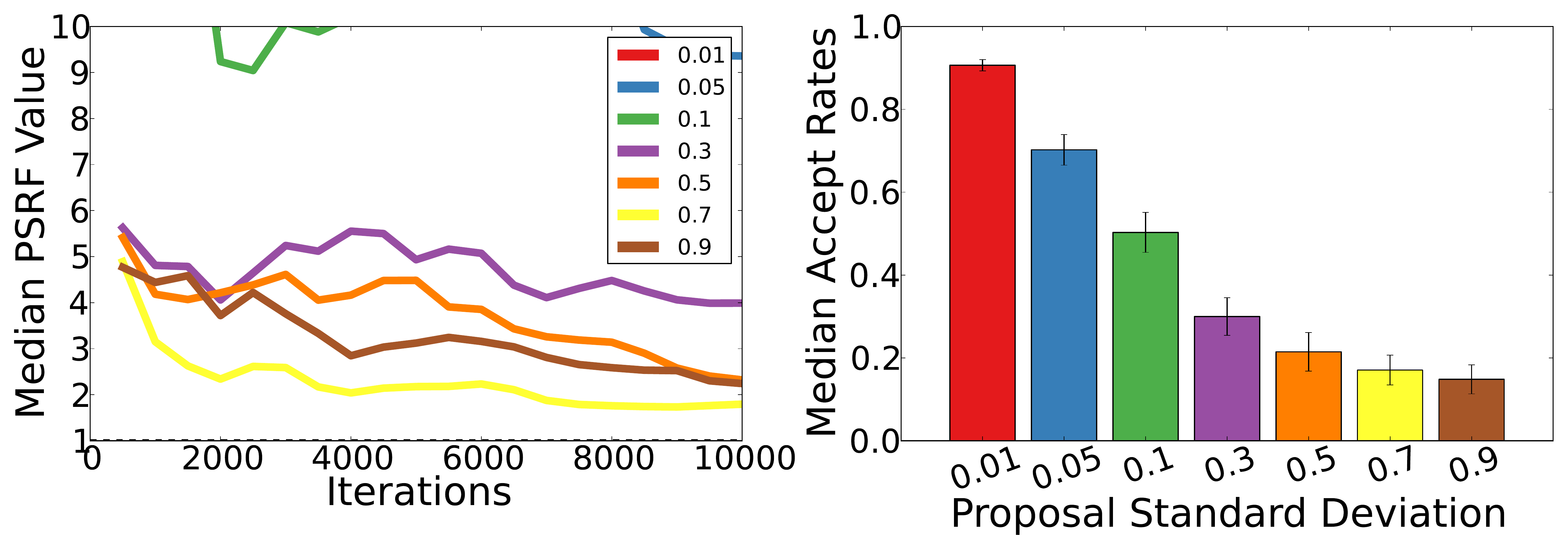} \label{fig:exp2_BMHWG} 
  }
\\
  \subfigure[\MHWG]{%
    \includegraphics[width=.48\textwidth]{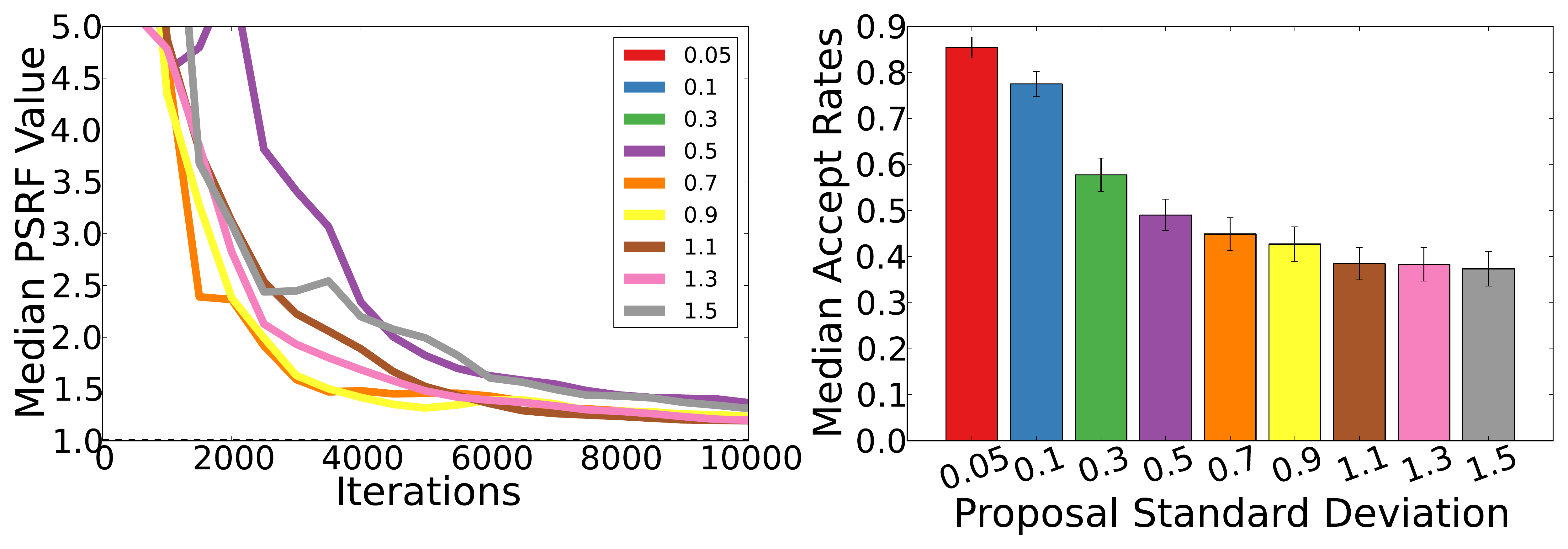} \label{fig:exp2_MHWG} 
  }
  \subfigure[\PT]{%
    \includegraphics[width=.48\textwidth]{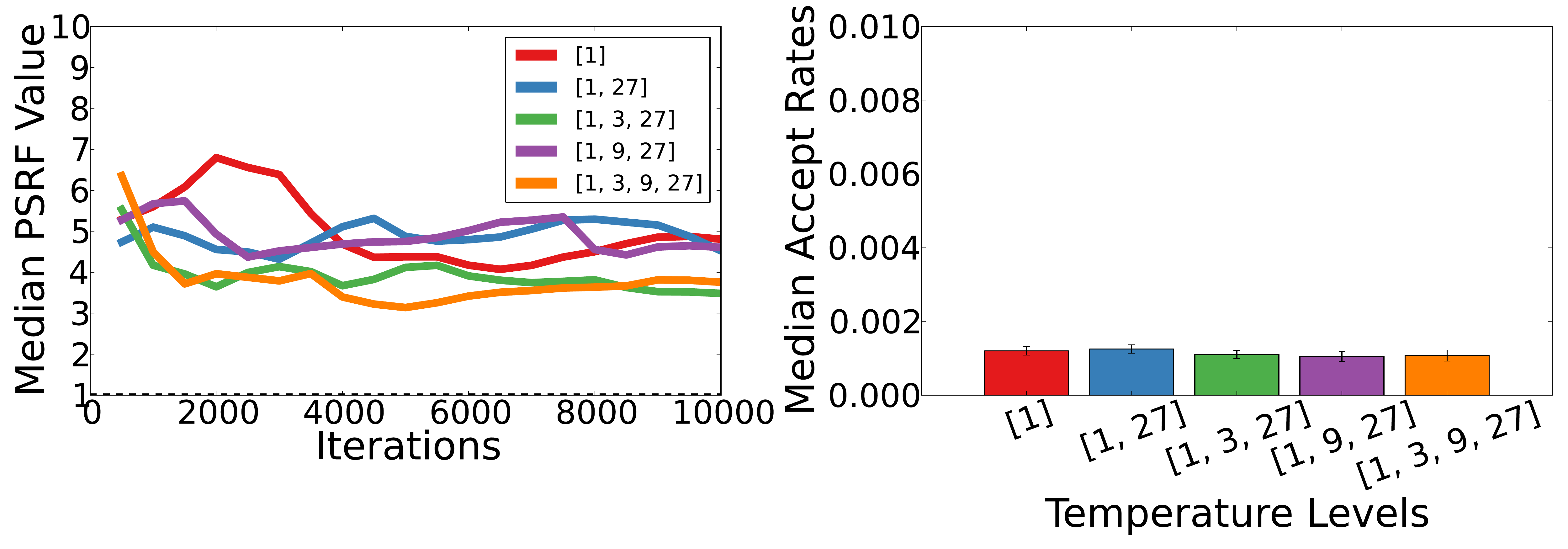} \label{fig:exp2_PT} 
  }
\\
  \subfigure[\INFBMHWG]{%
    \includegraphics[width=.5\textwidth]{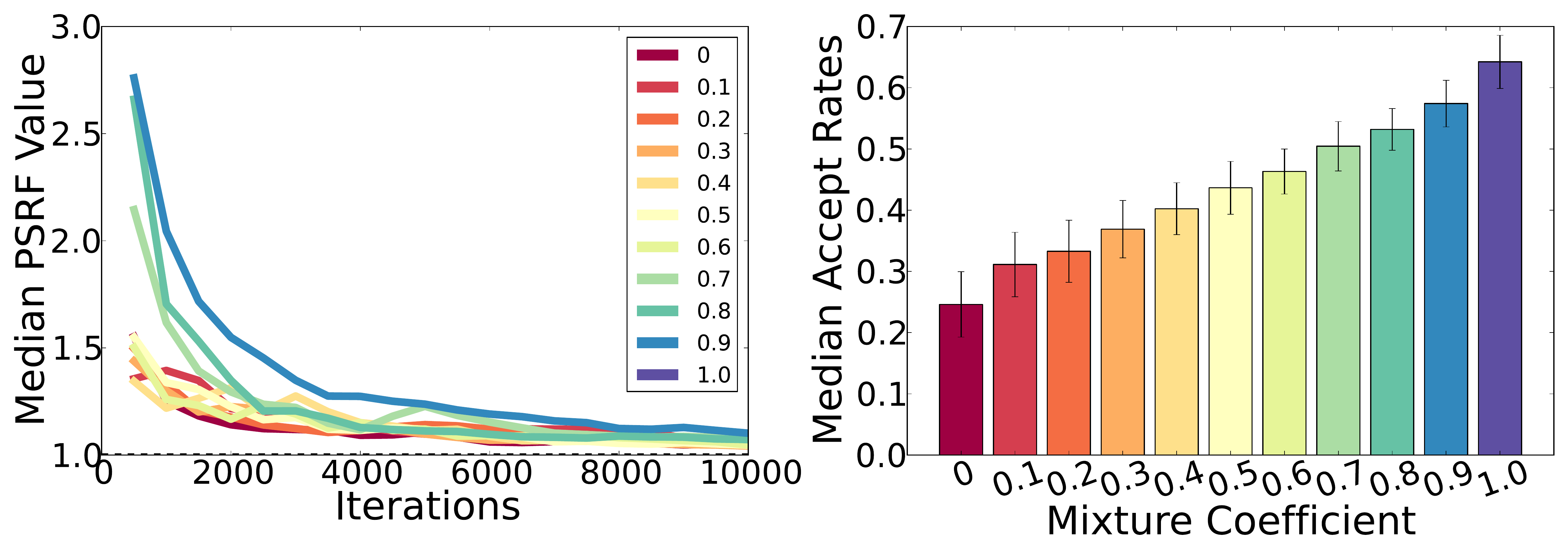} \label{fig:exp2_alpha} 
  }
  \caption{Results of the `Occluding Tiles' experiment. PRSF and
    Acceptance rates corresponding to various standard deviations of
    (a) \MH, (b) \BMHWG, (c) \MHWG,; (d) various temperature level
    combinations of \PT~sampling and; (e) various mixture coefficients
    of our informed \INFBMHWG sampling.}
\end{figure}

\section{Experiment: Occluding Tiles}

\subsection{Parameter Selection}

\paragraph{Metropolis Hastings (\MH)}

Figure~\ref{fig:exp2_MH} shows the results of 
\MH~sampling. Results show the poor convergence for all proposal
standard deviations and rapid decrease of AR with increasing standard
deviation. This is due to the high-dimensional nature of
problem. We selected a standard deviation of $1.1$.

\paragraph{Blocked Metropolis Hastings Within Gibbs (\BMHWG)}

The results of \BMHWG are shown in Figure~\ref{fig:exp2_BMHWG}. In
this sampler we update only one block of tile variables (of dimension
four) in each sampling step. Results show much better performance
compared to plain \MH. The optimal proposal standard deviation for
this sampler is $0.7$.

\paragraph{Metropolis Hastings Within Gibbs (\MHWG)}

Figure~\ref{fig:exp2_MHWG} shows the result of \MHWG sampling. This
sampler is better than \BMHWG and converges much more quickly. Here
a standard deviation of $0.9$ is found to be best.

\paragraph{Parallel Tempering (\PT)}

The Figure~\ref{fig:exp2_PT} shows the results of \PT sampling with various
temperature combinations. Results show no improvement in AR from plain
\MH sampling and again $[1,3,27]$ temperature levels are found to be optimal.

\paragraph{Effect of Mixture Coefficient in Informed Sampling (\INFBMHWG)}

Figure~\ref{fig:exp2_alpha} shows the effect of mixture
coefficient ($\alpha$) on the blocked informed sampling
\INFBMHWG. Since there is no significant different in PSRF values for
$0 \le \alpha \le 0.8$, we chose $0.8$ due to its high acceptance
rate.

\section{Experiment: Estimating Body Shape}

\subsection{Parameter Selection}
\paragraph{Metropolis Hastings (\MH)}

\begin{figure}[t]
\centering
  \subfigure[\MH]{%
    \includegraphics[width=.48\textwidth]{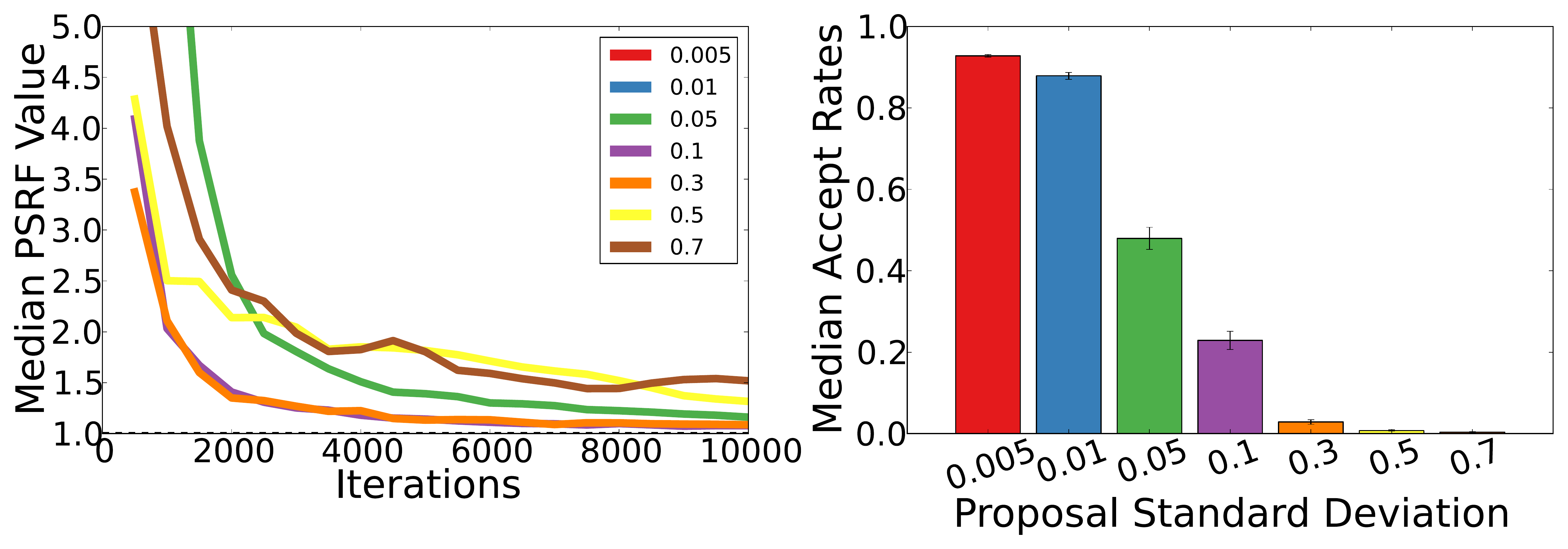} \label{fig:exp3_MH} 
  } 
  \subfigure[\MHWG]{%
    \includegraphics[width=.48\textwidth]{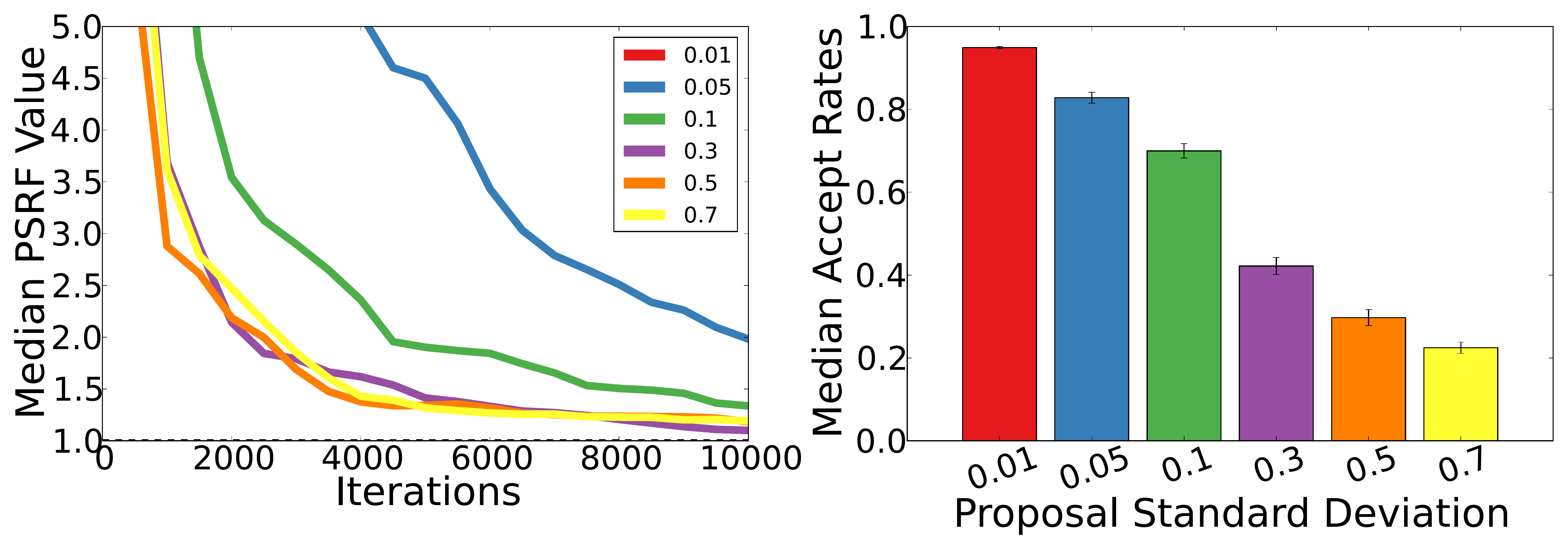} \label{fig:exp3_MHWG} 
  }
\\
  \subfigure[\PT]{%
    \includegraphics[width=.48\textwidth]{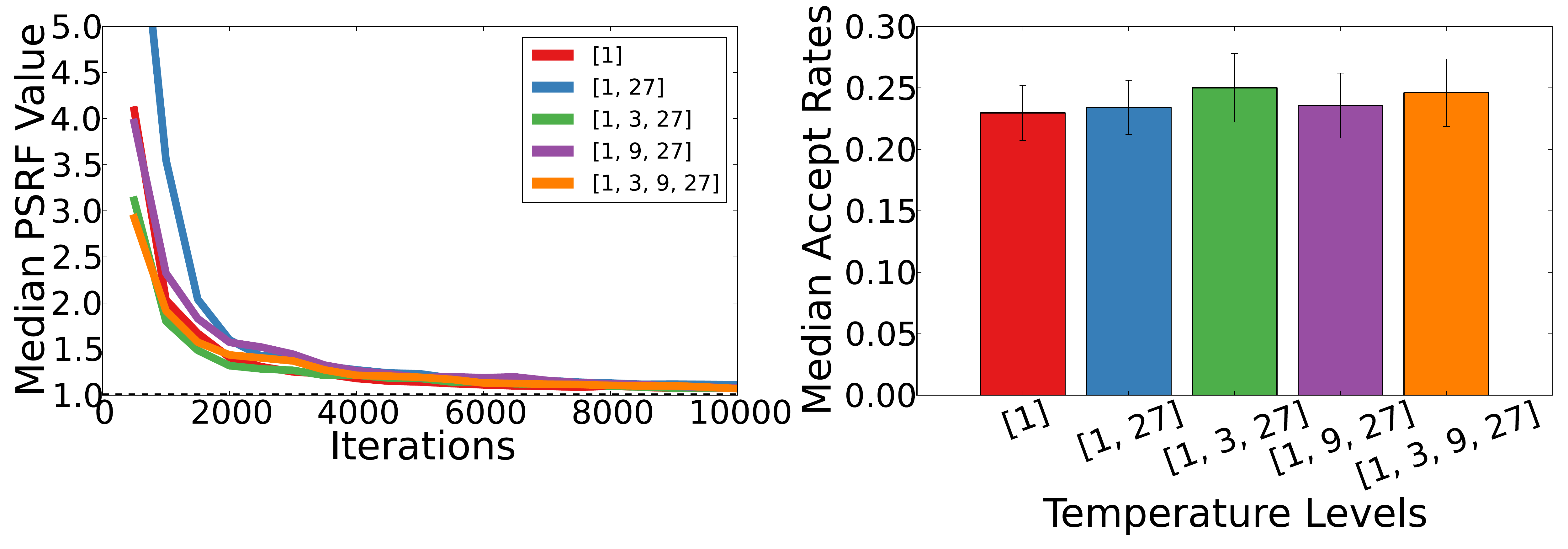} \label{fig:exp3_PT} 
  }
  \subfigure[\MIXLMH]{%
    \includegraphics[width=.48\textwidth]{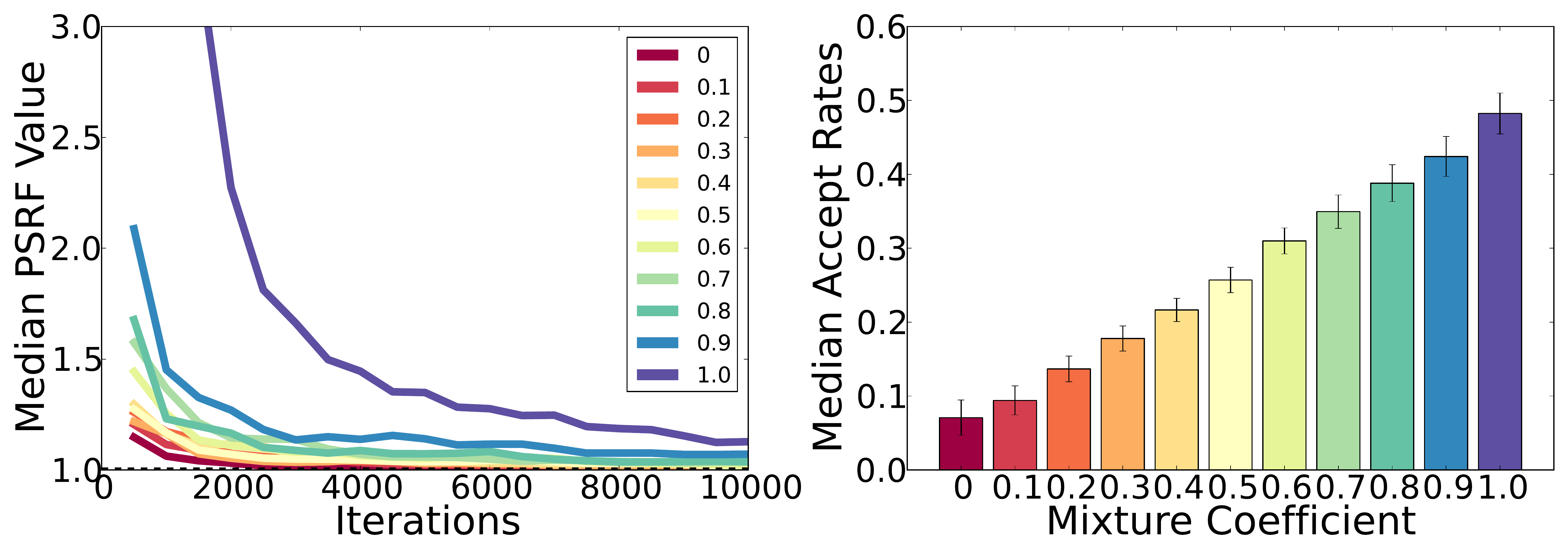} \label{fig:exp3_alpha} 
  }
\\
  \caption{Results of the `Body Shape Estimation' experiment. PRSFs and
    Acceptance rates corresponding to various standard deviations of
    (a) \MH, (b) \MHWG; (c) various temperature level combinations
    of \PT sampling and; (d) various mixture coefficients of the
    informed \MIXLMH sampling. }
\end{figure}

Figure~\ref{fig:exp3_MH} shows the result of \MH~sampling with various
proposal standard deviations. The value of $0.1$ is found to be
best.

\paragraph{Metropolis Hastings Within Gibbs (\MHWG)}

For \MHWG sampling we select $0.3$ proposal standard
deviation. Results are shown in Figure~\ref{fig:exp3_MHWG}.

\paragraph{Parallel Tempering (\PT)}

As before, results in Figure~\ref{fig:exp3_PT}, the temperature levels
were selected to be $[1,3,27]$ due its slightly higher AR.

\paragraph{Effect of Mixture Coefficient in Informed Sampling (\MIXLMH)}

Figure~\ref{fig:exp3_alpha} shows the effect of $\alpha$ on PSRF and
AR. Since there is no significant differences in PSRF values for $0 \le
\alpha \le 0.8$, we choose $0.8$.

\onecolumn\newpage\
\section{Results Overview}
\begin{figure*}[h]
\centering
  \subfigure[Results for: Estimating Camera Extrinsics]{%
    \includegraphics[width=0.9\textwidth]{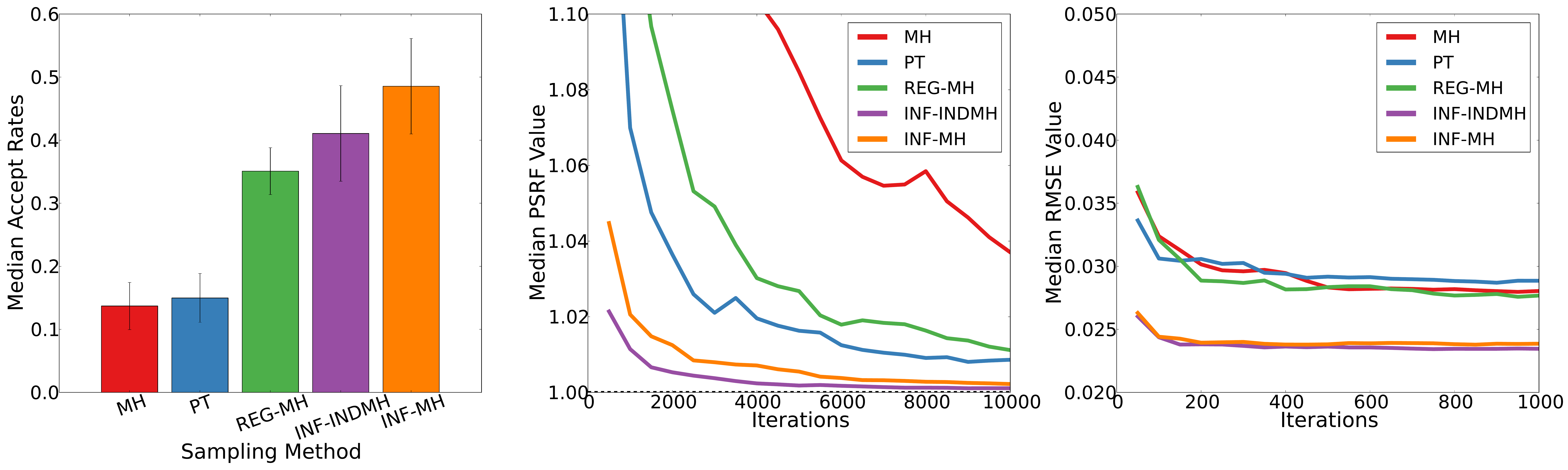} \label{fig:exp1_all} 
  }
  \subfigure[Results for: Occluding Tiles]{%
    \includegraphics[width=0.9\textwidth]{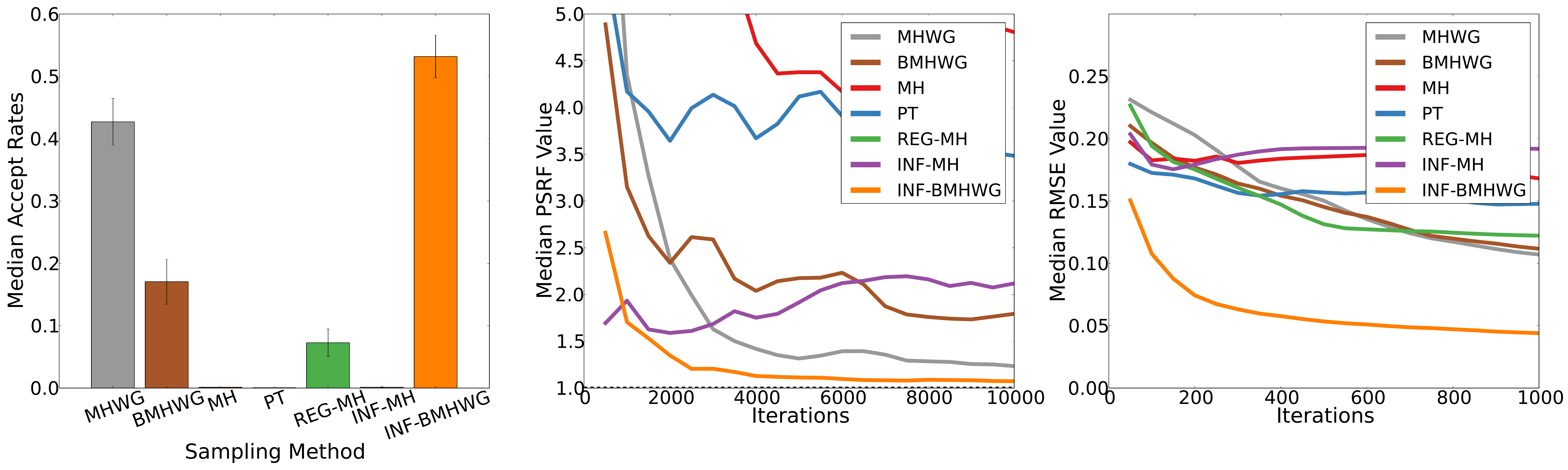} \label{fig:exp2_all} 
  }
  \subfigure[Results for: Estimating Body Shape]{%
    \includegraphics[width=0.9\textwidth]{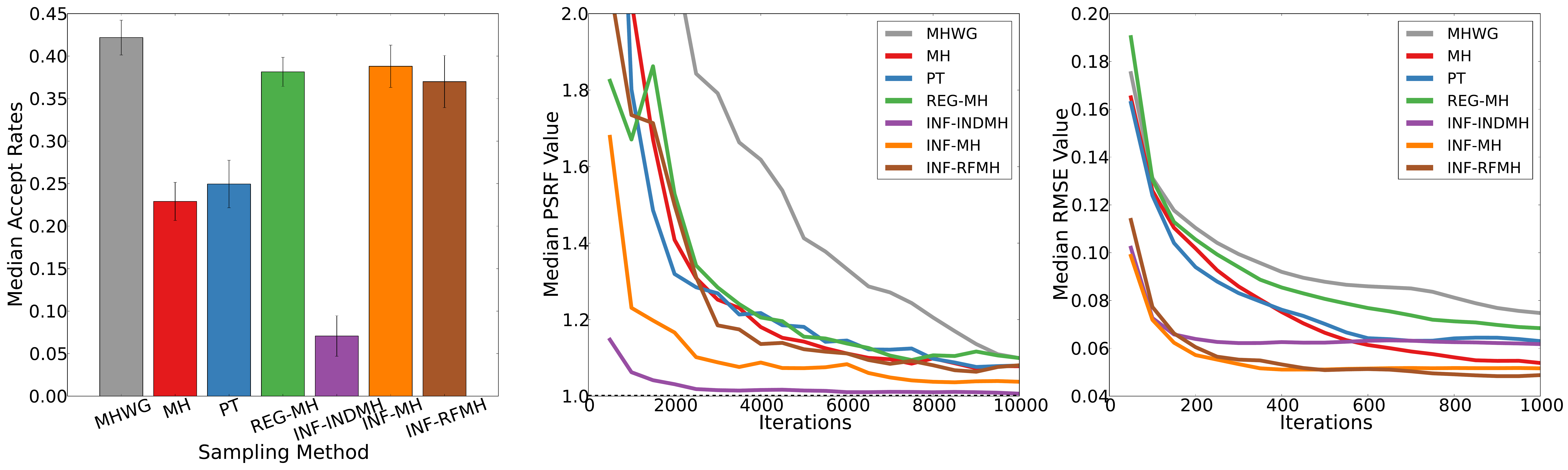} \label{fig:exp3_all} 
  }
  \caption{Summary of the statistics for the three experiments. Shown are
    for several baseline methods and the informed samplers the
    acceptance rates (left), PSRFs (middle), and RMSE values
    (right). All results are median results over multiple test
    examples. }
\end{figure*}

<

%

\end{document}